\documentclass{article}

\usepackage{PRIMEarxiv}

\usepackage[utf8]{inputenc} 
\usepackage[T1]{fontenc}    
\usepackage{hyperref}       
\usepackage{url}            
\usepackage{booktabs}       
\usepackage{amsfonts}       
\usepackage{nicefrac}       
\usepackage{microtype}      
\usepackage{lipsum}
\usepackage{fancyhdr}       
\usepackage{graphicx}       
\graphicspath{{media/}}     
\usepackage{subfigure}
\usepackage{multirow}
\usepackage{subcaption}
\usepackage{colortbl} 
\usepackage{makecell}
\usepackage{tabularx}
\usepackage{amsmath}
\usepackage[table]{xcolor}
\usepackage{ulem} 
\usepackage{xcolor} 
\usepackage{float} 

\newcommand{\figref}[1]{Fig.~\ref{#1}}%
\newcommand{\tabref}[1]{Tab.~\ref{#1}}%
\newcommand{\secref}[1]{Sec.~\ref{#1}}

\definecolor{wblue}{RGB}{106,205,243}
\definecolor{tblue}{RGB}{33,95,154}
\definecolor{tred}{RGB}{157,7,3}
\definecolor{tgreen}{RGB}{59,125,35}

\title{Lumina-Image 2.0: A Unified and Efficient Image Generative Framework
}

\author{Qi Qin$^{2}$\thanks{Equal Contribution}, ~Le Zhuo$^{1*}$, Yi Xin$^{1,4*}$, Ruoyi Du$^{1*}$, Zhen Li$^{3*}$, Bin Fu$^{1*}$, Yiting Lu$^{1*}$, Jiakang Yuan$^{1}$, Xinyue Li$^{1}$,   \\ \textbf{Dongyang Liu}$^{1,3}$,  \textbf{Xiangyang Zhu}$^{1}$, \textbf{Manyuan Zhang}$^{3}$, \textbf{Will Beddow}$^{6}$, \textbf{Erwann Millon}$^{6}$, \textbf{Victor Perez}$^{6}$,  \\ \textbf{Wenhai Wang}$^{1}$, \textbf{Conghui He}$^{1}$,  \textbf{Bo Zhang}$^{1}$, \textbf{Xiaohong Liu}$^{5}$, 
\textbf{Hongsheng Li}$^{3}$, \textbf{Yu Qiao}$^{1}$, \textbf{Chang Xu}$^{2}$, \textbf{Peng Gao}\textsuperscript{1}\thanks{Corresponding Authors} \thanks{Project Leader}\\ \\
  $^{1}$ Shanghai AI Laboratory, $^{2}$ The University of Sydney,  $^{3}$ The Chinese University of Hong Kong, \\ $^{4}$ Shanghai Innovation Institute, $^{5}$ Shanghai Jiao Tong University, $^{6}$ Krea AI\\
   \\
}

\begin{document}
\maketitle

\begin{figure*}[!ht]
    \centering
    \vspace{-0.15cm}
    \includegraphics[width=1.0\columnwidth]{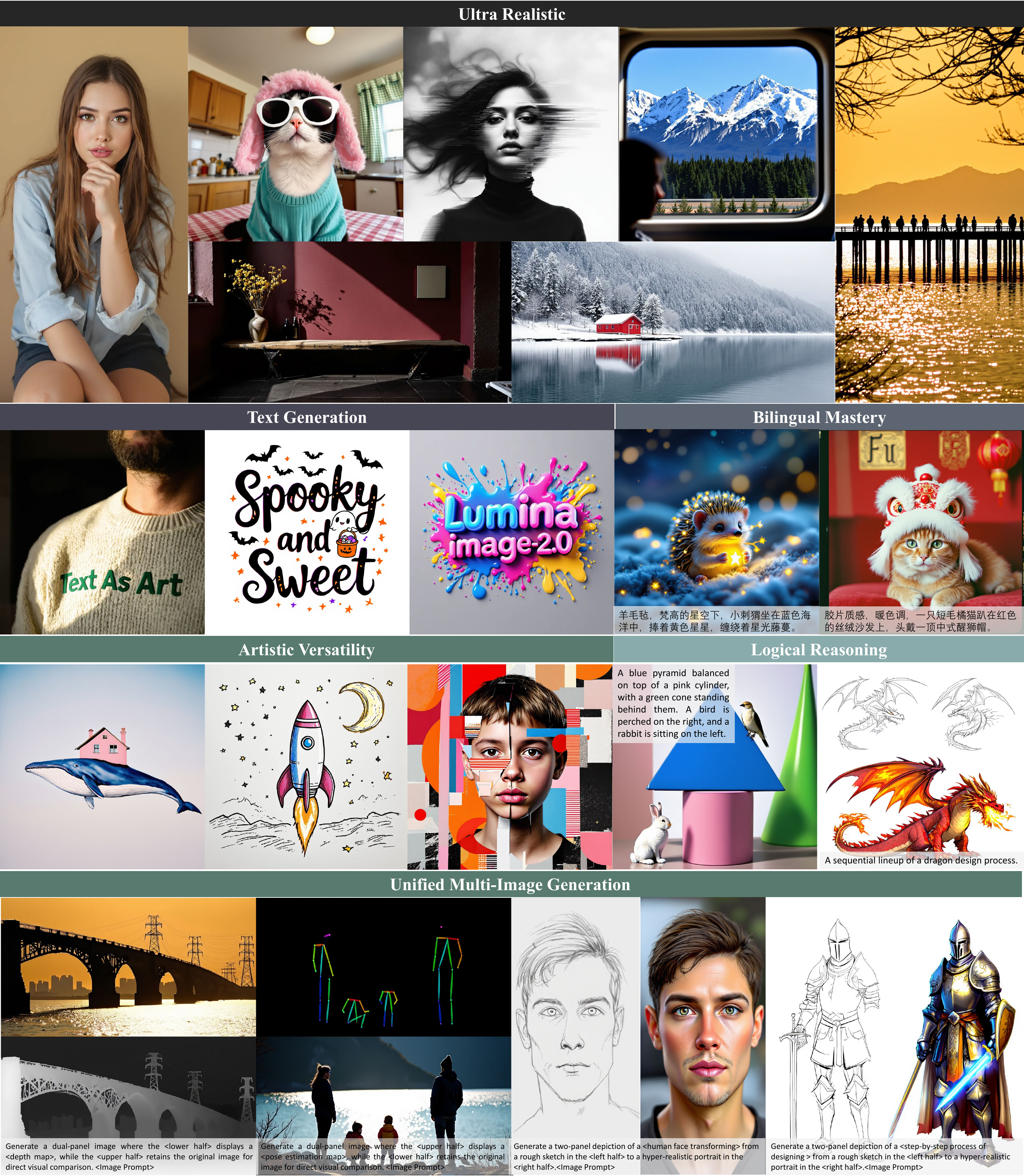}
    \vspace{-0.3cm}
    \caption{High-quality samples from our Lumina-Image 2.0, showcasing its capabilities in ultra-realistic, text generation, artistic versatility, bilingual mastery, logical reasoning, and unified multi-image generation.}
    \vspace{-0.4cm}
    \label{fig:demo}
\end{figure*}

\begin{abstract}
We introduce \textbf{Lumina-Image 2.0}, an advanced text-to-image generation framework that achieves significant progress compared to previous work, Lumina-Next.
Lumina-Image 2.0 is built upon two key principles: (1) \textit{Unification} – it adopts a unified architecture (Unified Next-DiT) that treats text and image tokens as a joint sequence, enabling natural cross-modal interactions and allowing seamless task expansion. 
Besides, since high-quality captioners can provide semantically well-aligned text-image training pairs, we introduce a unified captioning system, Unified Captioner (UniCap), specifically designed for T2I generation tasks. 
UniCap excels at generating comprehensive and accurate captions, accelerating convergence and enhancing prompt adherence. 
(2) \textit{Efficiency} – to improve the efficiency of our proposed model, we develop multi-stage progressive training strategies and introduce inference acceleration techniques without compromising image quality. 
Extensive evaluations on academic benchmarks and public text-to-image arenas show that Lumina-Image 2.0 delivers strong performances even with only 2.6B parameters, highlighting its scalability and design efficiency.
We have released our training details, code, and models at \url{https://github.com/Alpha-VLLM/Lumina-Image-2.0}. 
\end{abstract}


\section{Introduction}
\label{sec:intro}
Text-to-image (T2I) generative models have made significant strides over the past years. Notable open-source models~\cite{rombach2022high, betker2023improving, xie2024show, sun2024autoregressive, team2024chameleon, tang2024hart, wang2024emu3, xie2024sana,  flux2023, chen2025januspro} have shown significant improvements in both image fidelity and prompt adherence, thereby broadening the scope of their applicability in diverse downstream tasks~\cite{zhang2023adding,li2024photomaker,kawar2023imagic,lin2024pixwizard}. 
From these advancements, (1) the scalable text-conditional Diffusion  Transformer (DiT) architectures, and (2) the large-scale, high-quality text-image datasets are witnessed as the most important factors for developing text-conditional image generative models.

However, existing models still exhibit notable limitations in both aspects. 
First, many text-conditional Diffusion Transformers~\cite{chen2023pixart,xie2024sana,gao2024lumin-t2x,zhuo2024lumina,ma2025step,chen2025goku} continue to rely on cross-attention mechanisms to inject textual information. This paradigm treats text embeddings as fixed external features, thus limiting the efficiency of multimodal fusion and may even introduce uni-directional bias when using text embedding extracted from causal large language models~\cite{lidit}. Moreover, extending these models to new tasks often requires specific architecture designs~\cite{zhang2023adding,ye2023ip}. Second, although recent efforts~\cite{betker2023improving,chen2023pixart,li2024hunyuan} have highlighted the importance of collecting high-quality image captions, the lack of a dedicated captioning system tailored for T2I generation has resulted in inaccurate and insufficiently image captions for text-image paired training data. The limitations in both architecture and data quality constrain the expressiveness of text and visual representations, ultimately impairing the model’s ability to faithfully follow user instructions in generating high-quality images. 


Driven by the aforementioned challenges, we present {\bf Lumina-Image 2.0}, a unified and efficient T2I generative framework that comprises four key components: (1) a Unified Next-DiT model for generating images that faithfully aligned with the text input, (2) a Unified Captioner (UniCap) for producing high-quality text-image pairs, and a series of specific designs for (3) efficient training and (4) efficient inference. Specifically, to address architectural limitations, our {\bf Unified Next-DiT} model utilizes a joint self-attention mechanism, enabling our model to process both textual and visual tokens in a fully end-to-end manner, similar to decoder-only transformers in recent large language models~\cite{ChatGPT,jaech2024openai,dubey2024llama,touvron2023llama}. This design facilitates seamless multimodal interaction, allowing for the integration of additional multimodal tokens or specific prompt templates to extend the model’s capabilities without modifying the core architecture.
In response to the scarcity of high-quality textual descriptions in paired text-image data, we introduce {\bf Unified Captioner (UniCap)}, a unified captioning system specifically designed for T2I generation. UniCap excels at precisely understanding complex scenes, and generating comprehensive and coherent multilingual descriptions. 
Leveraging these capabilities, we employ UniCap to create multi-granularity, multi-dimensional textual descriptions that better align with the images.  
Furthermore, our experiments reveal that when combining the unified Next-DiT and UniCap for training, the text-to-image attention in transformer blocks dynamically adjusts its capacity based on the length of textual embeddings, behaving similarly to a dynamic feed-forward network. 
This observation motivates us to further enhance model capacity and performance by increasing the richness of textual descriptions without introducing additional parameters. 

Furthermore, both training and inference efficiency are crucial for model development and deployment. To perform  {\bf efficient training}, 
Lumina-Image 2.0 employs a multi-stage progressive training strategy with hierarchical high-quality data. The multi-domain system prompts and an auxiliary loss are further utilized to learn domain-specific knowledge and preserve low-frequency features, respectively. 
For {\bf efficient inference}, 
we adopt several advanced sampling techniques and verify that the integration of CFG-Renormalization (CFG-Renorm)~\cite{lin2024stiv} and CFG-Truncation (CFG-Trunc)~\cite{yi2024towards} can boost the sampling speed and maintain high sampling quality. 
Specifically, CFG-Renorm addresses the issue of over saturation at large classifier-free guidance (CFG) scales, while CFG-Trunc streamlines the inference process by eliminating redundant CFG computations, thereby enhancing both inference stability and speed. Additionally, we incorporate Flow-DPM-Solver~\cite{xie2024sana} and TeaCache~\cite{liu2024timestep} to further optimize inference speed.

We evaluate Lumina-Image 2.0 on publicly available benchmarks, including DPG~\cite{hu2024ella}, GenEval~\cite{ghosh2024geneval}, and T2I-CompBench~\cite{huang2023t2i}. Considering the limitation of current academic benchmarks in comprehensively evaluating T2I models, we report the ELO rankings of Lumina-Image 2.0 on several online T2I arenas, which is evaluated by human annotators. 
Our experimental results consistently demonstrate that Lumina-Image 2.0 achieves significant improvements over previous model (Lumina-Next~\cite{zhuo2024lumina}). 
We release the complete training details, code, and models to facilitate the full reproduction of Lumina-Image 2.0.

\section{Related Work}
Recent advancements in text-to-image generation have been remarkable. Diffusion-based models have progressively transitioned from U-Net architectures~\cite{podell2023sdxl} to Diffusion Transformers~\cite{dit}, as demonstrated by models such as PixArt~\cite{chen2023pixart,chen2024pixart}, FLUX~\cite{flux2023}, SD3~\cite{esser2024scaling}, Lumina~\cite{gao2024lumin-t2x,zhuo2024lumina}, and SANA~\cite{xie2024sana}. 
These Diffusion Transformers exhibit superior scalability and are progressively evolving toward a unified multimodal representation~\cite{xiao2024omnigen}. 
Regarding text encoders, early approaches~\cite{rombach2022high} employed CLIP \cite{radford2021learning}, while subsequent works~\cite{li2024hunyuan,esser2024scaling,flux2023} additionally adopted T5-XXL~\cite{2020t5}. 
More recently, SANA~\cite{xie2024sana}, Lumina~\cite{zhuo2024lumina,gao2024lumin-t2x} and our Lumina-Image 2.0 have incorporated Gemma~\cite{gemma} as the text encoder. 
Furthermore, the latest models leverage flow-based parameterizations~\cite{liu2022flow,lipman2022flow}, which enhance both training and inference efficiency compared to conventional diffusion methods. 
In parallel, a range of advanced autoregressive and hybrid text-to-image models have emerged~\cite{liu2024lumina,xie2024show,tang2024hart,han2024infinity,chen2025januspro,wang2024emu3}, achieving performance on par with their diffusion-based counterparts. 
However, the sampling speed of these autoregressive models remains significantly slower than that of diffusion-based approaches, posing a critical challenge for their practical deployment.

Meanwhile, the advancement of text-to-image models has been significantly shaped by the evolution of vision-language models (VLMs)~\cite{liu2024visual,chen2024sharegpt4v,chen2024internvl,liu2024sphinx}, where the quality of image captions plays a critical role in both model performance~\cite{betker2023improving,esser2024scaling}. 
Currently, the most commonly employed image captioners in text-to-image research include LLaVA~\cite{liu2023visual},  CogVLM~\cite{wang2023cogvlm}, ShareGPT-4~\cite{chen2024sharegpt4v}, and Qwen-VL~\cite{bai2023qwen,Qwen2.5-VL,Qwen2-VL}, all of which are general-purpose vision-language models (VLMs). 
However, there is a significant lack of research focused on developing captioner models specifically tailored for the text-to-image task, which may impede the further advancement of text-to-image models. 

\section{Revisiting Lumina-Next}

\textbf{Model Architecture.} 
Lumina-Next \cite{zhuo2024lumina} introduces Next-DiT, a scalable flow-based diffusion transformer, as its core architecture. Building upon the original diffusion transformer~\cite{dit}, Next-DiT employs sandwich normalization and query-key normalization~\cite{dehghani2023scaling} to enhance training stability, and leverages 2D Rotary Positional Encoding~\cite{su2021roformer} to encode positional information of images. 
For text-to-image generation, Next-DiT utilizes zero-initialized gated cross-attention to inject text embeddings extracted by Gemma~\cite{gemma}.

\textbf{Training and Inference Strategy.} Lumina-Next is trained on approximately 20M synthetic text-image pairs, with image captions generated using user prompts and VLM models. During training, Lumina-Next employs a multi-stage progressive training approach, similar to recent text-to-image models~\cite{chen2023pixart,gao2024lumin-t2x}. This strategy involves sequential training at 256, 512, and 1024 resolutions, enabling the model to progressively capture both low-frequency and high-frequency information from images. During sampling, Lumina-Next introduces two time schedules tailored for flow models to minimize the ODE truncation errors, and it supports both first-order and higher-order solvers, such as Euler and Midpoint solvers. 

\textbf{Naive Data Scaling.} 
\label{scaling_data}
Inspired by the data scaling paradigm of large-scale models~\cite{achiam2023gpt,jaech2024openai,team2023gemini,esser2024scaling}, we believe that the performance gap in Lumina-Next primarily stems from insufficient training data. Therefore, we scaled Lumina-Next’s dataset from 20M to 200M samples. This expanded dataset encompasses diverse real and synthetic data processed by the same cleaning and annotation pipeline. We retain the same model architecture and training strategy in Lumina-Next. We observed that the model performance shows considerable improvement across various academic metrics compared to Lumina-Next. For example, on the DPG benchmark \cite{hu2024ella}, the performance improved from 75.66 to 85.80 after data scaling. This demonstrates the effectiveness of Next-DiT as a robust framework for scalable image generation.



\section{Lumina-Image 2.0}

\subsection{Framework Overview}
Lumina-Image 2.0 establishes a unified and efficient framework by integrating Unified Next-DiT, Unified Captioner (UniCap), and a set of efficient training and inference strategies. 
The overall pipeline is illustrated in~\figref{fig:pipeline},  
and we apply a custom filtering pipeline \cite{chen2023pixart,kong2024hunyuanvideo} to select high-quality training images. 
To improve text quality, our UniCap re-captions the training data to generate accurate and detailed textual descriptions at multiple levels of granularity. 
The resulting high-quality image-text pairs are organized into a hierarchical training dataset, which is subsequently used to optimize the Unified Next-DiT model using our proposed training strategies. 
Finally, several inference strategies are further introduced to efficiently generate high-quality images. 

\subsection{Unified Next-DiT}
After revisiting the architecture of Next-DiT, we observe that zero-initialized gated cross-attention for integrating text embedding limits the capability of text-image alignment and also requires additional architecture modification when adapting to new tasks. 
Therefore, we propose Unified Next-DiT, a unified text-to-image model that treats text and images as a unified sequence to perform joint attention inspired by recent advances in unified multimodal learning~\cite{xiao2024omnigen}.

\begin{figure*}[!t]
    \centering
\includegraphics[width=0.85\columnwidth]{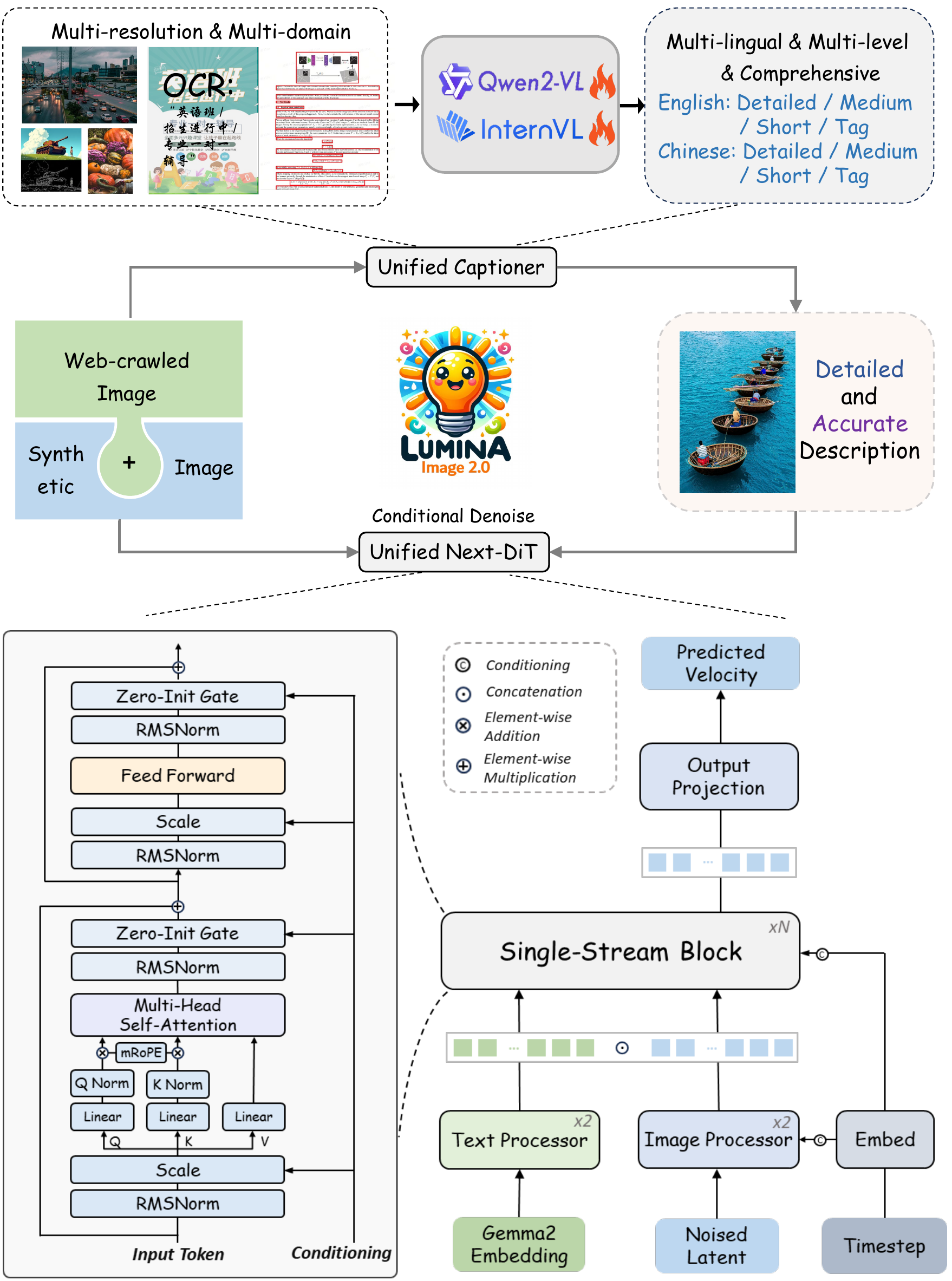}
\vspace{0.2cm}
    \caption{Overview of Lumina-Image 2.0, which consists of Unified Captioner and Unified Next-DiT. The Unified Captioner re-captions web-crawled and synthetic images to construct hierarchical text-image pairs, which are then used to optimize Unified Next-DiT with our efficient training strategy.} 
    \label{fig:pipeline}
    
\end{figure*}

\noindent \textbf{Architecture of Unified Next-DiT.}
Next-DiT employs Gemma~\cite{gemma} as the text encoder, whose text embeddings exhibit unidirectional positional bias~\cite{lidit} caused by the causal self-attention in the large language model. During generation, the biased text embeddings are fixed and sparsely injected to the transformer block via zero-initialized gated cross-attention. Therefore, we remove all zero-initialized gated cross-attention in Next-DiT. Instead, we leverage a unified single-stream block that fuses caption embeddings and noised latent by concatenating them and performing joint self-attention, which facilitates more effective text–image interaction and task expansion. 
As illustrated in~\figref{fig:pipeline}, our single-stream blocks build upon the original DiT block with the addition of sandwich normalization and query-key normalization to ensure stable training. 
The Multimodal-RoPE~\cite{qwen2vl} (mRoPE) is employed to jointly model text–image sequences in a unified manner, which encodes the text length as well as the image’s height and width into three dimensions. 
Moreover, we further observe that textual and visual features at the input level exhibit a considerable gap. 
To address this issue, we introduce text and image processors prior to the single-stream blocks.
These processors with similar but lightweight single-stream blocks facilitate intra-modal information exchange and mitigate the gap between modalities.
Since caption embeddings are fixed for all timesteps, the text processor does not incorporate timestep conditioning.

\begin{figure*}[!h]
    \centering
    \includegraphics[scale=0.58]{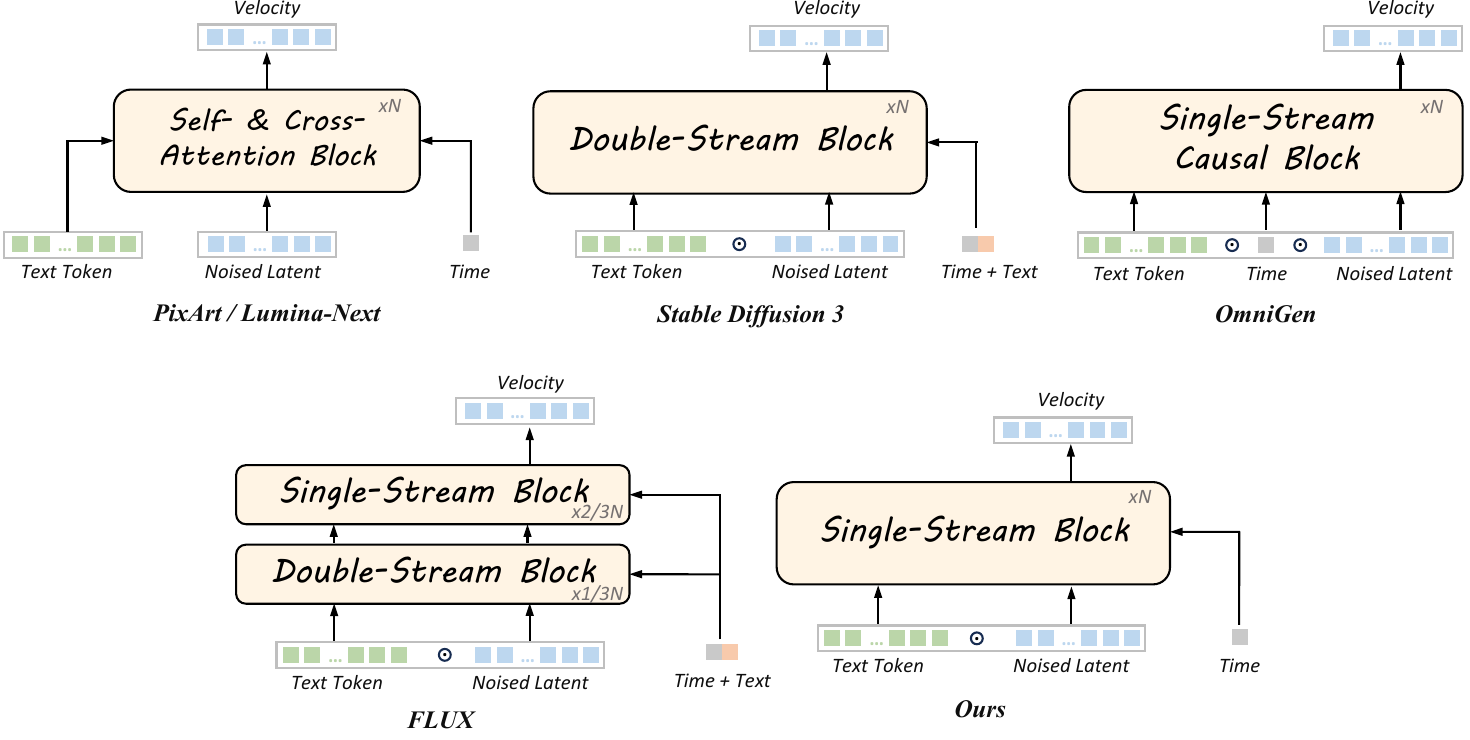}
    \caption{We compare the Diffusion Transformer architectures between our Unified Next-DiT, and PixArt~\cite{chen2023pixart}, Lumina-Next~\cite{zhuo2024lumina}, Stable Diffusion 3~\cite{esser2024scaling}, OmniGen~\cite{xiao2024omnigen} and FLUX~\cite{flux2023}.}
    \label{fig:block-comparation}
\end{figure*}

\noindent \textbf{Comparison with previous architectures.}
As illustrated in \figref{fig:block-comparation}, we compare the architecture of Unified Next-DiT with mainstream Diffusion Transformers. PixArt~\cite{chen2023pixart} and Lumina-Next~\cite{zhuo2024lumina} employ an additional cross-attention block after self-attention to inject fixed text embeddings, whereas our model adopts a single, unified attention module that jointly handles both text and noisy latent. Compared with the MMDiT architecture used in SD3~\cite{esser2024scaling} and FLUX~\cite{flux2023}, the key difference is that  MMDiT employs the double-stream blocks, allocating extensive and separate parameters for text and image sequences. In contrast, our method is designed from a more unified perspective, utilizing a single set of parameters to simultaneously model both the text and image sequences. 
Our model shares similarities with OmniGen~\cite{xiao2024omnigen}, which introduces a single-stream causal DiT architecture for unified image generation. 
In pursuit of unifying the transformer architecture with auto-regressive models, OmniGen removes adaptive Layer Normalization (adaLN) and applies causal self-attention initialized from a large language model. However, adaLN is considered essential for Diffusion Transformers~\cite{dit}, and initializing from a language model may introduce conflicts with the knowledge for image generation.

\subsection{Unified Captioner} 


Due to the crucial role of image captions in enhancing model performance~\cite{betker2023improving}, using out-of-box pre-trained Vision Language Models (VLMs) for image recaptioning has been standard practice in previous literatures~\cite{li2024hunyuan,esser2024scaling,chen2023pixart,chen2024pixart,gao2024lumin-t2x,zhuo2024lumina}. However, these VLMs exhibit several limitations, including single-granularity descriptions, domain biases, and fixed low-resolution inputs, which result in suboptimal caption quality and a noticeable gap from real-world user prompts. 
To address these limitations and construct high-quality text-image datasets, we develop Unified Captioner (UniCap), a captioning system that unifies diverse visual inputs and provides multi-granularity, multi-perspective, and multi-lingual high-quality textual descriptions. In addition, we also introduce a unified perspective to make the caption-driven model capacity more interpretable.

\noindent \textbf{Unifying Textual Description.}
To enable Lumina-Image 2.0 to handle diverse prompts -- ranging from multi-granularity, multi-perspective, and multilingual descriptions -- we train UniCap to deliver all types of descriptions to achieve unified image recaptioning. 
In particular, our approach comprises three key components: (1) For multi-granularity descriptions, we begin by carefully prompting GPT-4o~\cite{achiam2023gpt} to generate highly detailed descriptions. Then we employ open-source large language models (LLMs) to simultaneously summarize detailed captions into medium, short, and tag-based descriptions for captioner training, which enables UniCap to deliver captions of multiple granularities while retaining essential information, as shown in \figref{fig:captioner_1} and \figref{fig:captioner_2}.
(2) For multi-perspective descriptions, we include image style descriptions, main object descriptions, all-object descriptions, object attribute descriptions, and spatial relationship descriptions, ensuring comprehensive coverage of visual elements, attributes, spatial structures, and stylistic nuances. 
(3) For multi-lingual descriptions, we utilize bilingual large language models to translate captions into Chinese, enabling UniCap to generate bilingual captions simultaneously.
Surprisingly, although UniCap only captions all data in English and Chinese for Lumina-Image 2.0 training, the model benefits from Gemma's multilingual capabilities and emerges with the understanding of other languages, thereby expanding its accessibility to a broader user base (see \figref{fig:multilingual}).

\noindent \textbf{Unifying Visual Understanding.}
Existing VLMs struggle with processing images from diverse domains and open-world scenarios, and are limited to low-resolution inputs, making it difficult to capture fine-grained details of images.
To address this issue, we train UniCap with a caption dataset that encompasses a wide range of visual content, including natural images, web-crawled images, photographs, synthetic images, multi-image documents, infographics, OCR-related images, and multilingual content, ensuring comprehensive domain coverage and conceptual diversity.
Besides, unlike LLaVA~\cite{liu2024visual} and ShareGPT4V~\cite{chen2024sharegpt4v}, which resize images of varying scales and aspect ratios to a fixed low-resolution format, our UniCap processes images at their native scale in a unified manner. This approach yields more accurate and detailed captions, significantly reduces hallucinations, and improves OCR recognition. 
This strategy has been widely adopted by recent VLMs, including SPHINX-X~\cite{liu2024sphinx}, InternVL~\cite{chen2024internvl}, and XComposer~\cite{zhang2023internlm}.

Furthermore, inspired by the concept of specialized generalist intelligence (SGI)~\cite{zhang2024towards}, where AI systems excel in specialized tasks while maintaining broad general abilities, we aim for Lumina-Image 2.0 to not only showcase powerful text-conditioned generation capabilities but also serve as a unified interface for diverse visual generation tasks.
To this end, we collect annotations from various visual tasks, such as depth maps, pose maps, canny maps, and sketches. We then concatenate them with paired images to form composite grids and leverage template captions (please refer to the last row of \figref{fig:demo}) to effectively describe the underlying logical process. 
This unified approach enables Lumina-Image 2.0 to handle advanced tasks beyond text-to-image generation, thereby laying the foundation for potential downstream applications. 
\label{sec:cap_ffn}

\label{sec:cap_ffn}
\begin{figure}[ht]
    \centering
    \includegraphics[width=0.6\columnwidth]{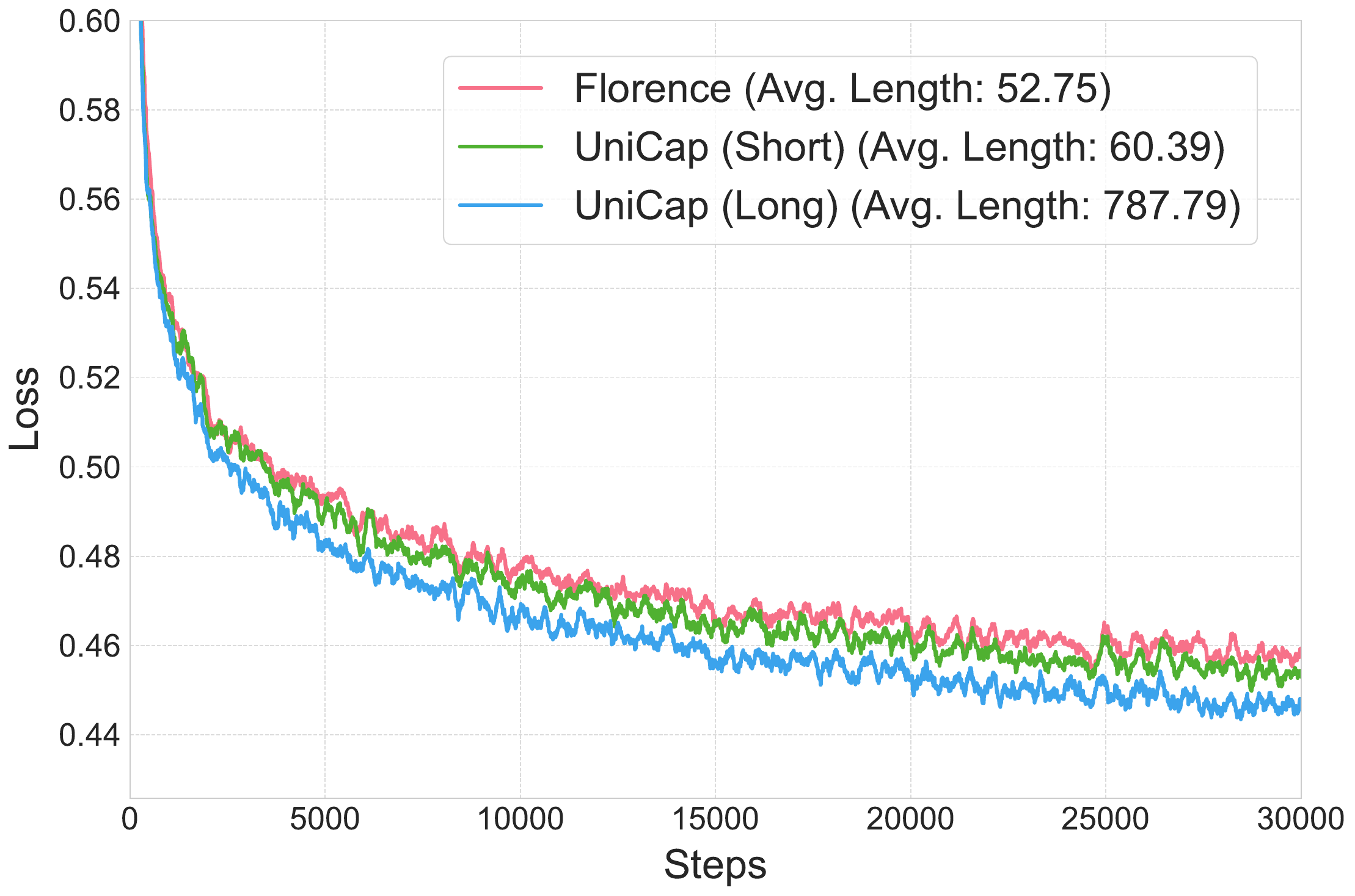}
    \caption{The training loss curve with respect to captions with different lengths. The “Avg. Length” represents the average character number.}
    \label{fig:loss_curve}
\end{figure}

\begin{figure}[t]
    \centering
    \includegraphics[width=0.85\columnwidth]{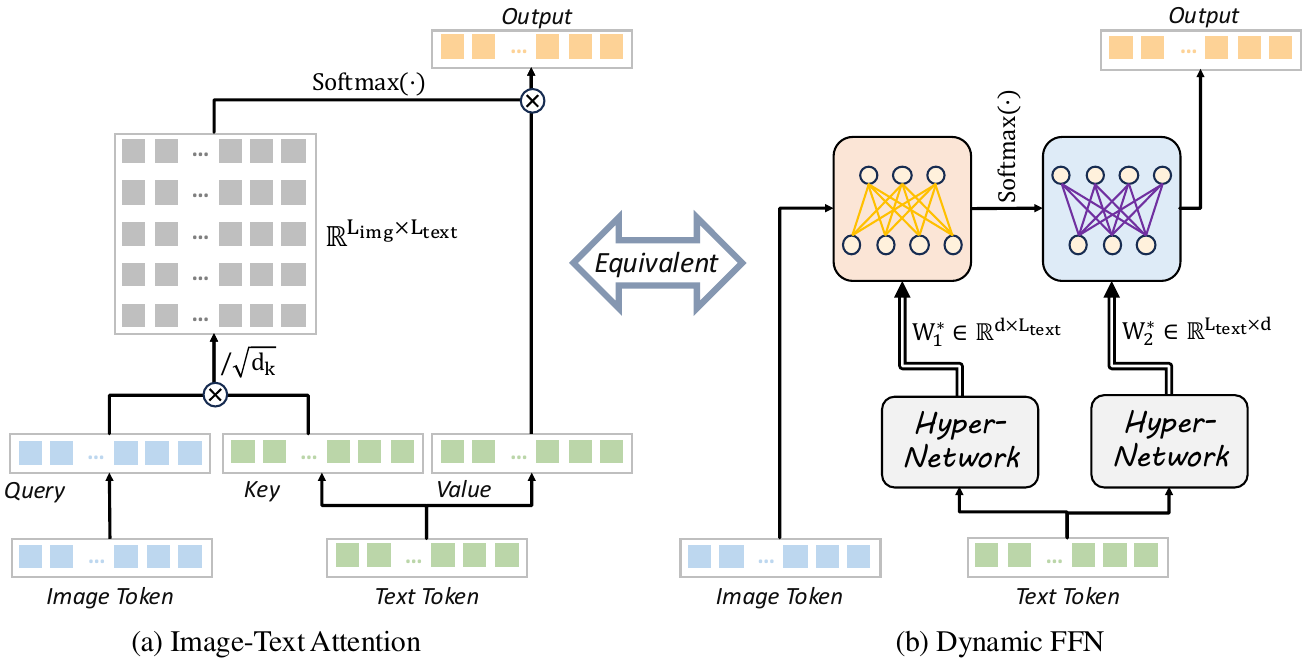}
    \caption{Illustration of reformulating the Image-Text Attention as an FFN generated by a hyper-network, with its weights and hidden dimensions dynamically determined by the input text token.}
    \label{fig:dynamic_ffn}
\end{figure}

\noindent \textbf{A Unified Perspective on Caption-Driven Model Capacity.}
The importance of detailed image captions has been witnessed for scaling up diffusion models~\cite{betker2023improving,esser2024scaling}.
During the training of Lumina-Image 2.0, we specifically observed that both the length and quality of image captions direct influence the model's convergence speed. As shown in \figref{fig:loss_curve}, we train the model using three versions of image captions: (1) short captions generated by Florence~\cite{xiao2024florence}, (2) short but precise captions generated by our UniCap, and (3) long and detailed captions from UniCap. 
We observe that as captions become more precise and detailed, the model's convergence speed significantly improved.
During the inference phase, it is also commonly recognized by previous works~\cite{kong2024hunyuanvideo,li2024hunyuan} that longer captions often lead to better generation results. These observations motivate us to rethink the role of caption embeddings in text-to-image generation.
To further analyze the impact of image caption, in this paper, we provide an interpretable perspective on this phenomenon -- the attention operation between texts and images can be viewed as a dynamic feed-forward network (FFN), where the choice of caption embeddings governs its effective knowledge integration and representational capacity. 

Generally, the FFN layer of a transformer can be interpreted as a key-value memory that encapsulates the general knowledge acquired by the model~\cite{geva2020transformer}, which can even be manually constructed without the need for training~\cite{zhang2021tip}. 
It also has been shown that the FFN layer can be effectively substituted by self-attention with persistent memory~\cite{sukhbaatar2019augmenting}. 
Motivated by these findings, we further explore the relationship between the text-to-image attention and the FFN mechanism.
Note that the term ``text-to-image attention'' encompasses both the independent cross-attention used in Next-DiT~\cite{zhuo2024lumina} and PixArt~\cite{chen2024pixart}, as well as the image-text interaction component in models performing joint self-attention (e.g., our Unified Next-DiT).

Given a sequence of image tokens \(X \in \mathbb{R}^{L_\text{img} \times d}\) and a sequence of text tokens \(Y \in \mathbb{R}^{L_\text{text} \times d}\). An ordinary image-text attention can be equivalently rewritten in the form of FFN as follows:
\begin{align}
\text{Attn}(X, Y) &= \sigma\!\bigl(X\,W_{1}(Y)\bigr)\;W_{2}(Y),
\end{align}
where $\sigma(\cdot)$ denotes the Softmax function,
and the two ``weight matrices'' are conditioned on the text embeddings \(Y\):
\begin{align}
W_{1}(Y) &= \frac{\,W_{Q}\,\bigl(Y\,W_{K}\bigr)^{T}}{\sqrt{d_{k}}} 
\quad \in \mathbb{R}^{d \times L_\text{text}},\\
W_{2}(Y) &= Y\,W_{V} 
\quad \in \mathbb{R}^{L_\text{text} \times d},
\end{align}
where $W_Q$, $W_K$, and $W_V$ are weight matrices for query, key, and value, respectively, and $d_k$ is the dimension of query / key.  Notably, the hidden dimension between $W_{1}(Y)$ and $W_{2}(Y)$ changes dynamically with the context length as \( L_\text{text}\). Under this formulation, the text-to-image attention computation can be viewed as an FFN whose parameters are generated by a hyper-network, with \textit{dynamic weights} and \textit{dynamic hidden size}.  Specifically, the conditional information (\textit{i.e.}, the text) is encoded to form the dynamic weights, while the hidden size \(\!L_\text{text}\) will adjust the capacity of this FFN-like module via its length. 

From this perspective, we reach an interesting conclusion -- 
increasing caption length effectively serves as a controllable means of scaling up model parameters. This insight suggests that the capacity of the model in both training and inference can be modulated simply by adjusting the length of the caption, which can lead to improved knowledge learning and overall performance. These findings are consistent with recent trends in existing work~\cite{ma2025inference,xie2025sana} and highlight promising directions such as inference-time scaling~\cite{snell2024scaling}.

\subsection{Efficient Training}
Lumina-Image 2.0 introduces an efficient training framework that integrates multi-stage progressive training, a hierarchical high-quality dataset, multi-domain system prompts, and auxiliary loss. These strategies improve image quality and detail refinement while accelerating convergence. 

\noindent \textbf{Multi-Stage Progressive Training.} Unlike prior approaches \cite{chen2023pixart,podell2023sdxl,zhuo2024lumina,xiao2024omnigen} that optimize generative models over three progressive resolution stages, we skip the intermediate  512 resolution stage and introduce an additional high-quality tuning phase. 
This results in a three-stage progressive training pipeline: a low-resolution phase (256 resolution), a high-resolution phase (1024 resolution), and a high-quality tuning phase (1024 resolution).
In the low-resolution phase, Lumina-Image 2.0 focuses on learning global and low-frequency information, such as domain knowledge, object relationships, and structural patterns. The subsequent two phases first transfer this knowledge to higher resolutions and further enhance fine-grained visual details. 

\noindent \textbf{Hierarchical High-quality Data.} In contrast to Lumina-Next, which utilizes a fixed dataset in all training stages, we construct a hierarchical dataset by filtering images based on image quality criteria (e.g., aesthetic) at different stages. 
Specifically, we begin with a dataset of 110M samples. For the low-resolution training stage, we select 100M samples. The remaining 10M samples, containing relatively higher-quality data,  are then used in the high-resolution phase. From these, we further curate a subset of the highest-quality 1M samples for the final fine-tuning stage.

\begin{table}[ht]
\centering
\caption{Prompt template for Lumina-Image 2.0. \textcolor{tgreen}{<Image Prompt>} will be replaced with the user specific image description. \textcolor{tblue}{<lower half>} and \textcolor{tblue}{<upper half>} will be replaced with the specific spatial relationships. \textcolor{orange}{<depth map>} will be replaced with the target image type.}
\renewcommand{\arraystretch}{1.7}  
\begin{tabular}{>{\centering\arraybackslash}m{2cm} m{10cm}}  
\hline
\textbf{Template A} & You are an assistant designed to generate high-quality images based on user prompts. \textcolor{tred}{ <Prompt Start> } \textcolor{tgreen}{<Image Prompt>} \\
\hline
\textbf{Template B} & You are an assistant designed to generate superior images with the superior degree of image-text alignment based on textual prompts or user prompts. \textcolor{tred}{ <Prompt Start> } \textcolor{tgreen}{<Image Prompt>} \\
\hline
\textbf{Template C} & Generate a dual-panel image where the \textcolor{tblue}{<lower half>} displays a \textcolor{orange}{<depth map>}, while the \textcolor{tblue}{<upper half>} retains the original image for direct visual comparison. \textcolor{tred}{ <Prompt Start> } \textcolor{tgreen}{<Image Prompt>} \\
\hline
\label{tab:system_prompt}
\end{tabular}
\end{table}

\noindent \textbf{Multi-domain System Prompt.}  
We collect training data from diverse domains, including high-aesthetic synthetic data, as well as photorealistic real-world data. 
However, there is a substantial domain gap between these datasets, often resulting in slower convergence and difficulties in learning domain-specific knowledge. 
Motivated by ChatGPT \cite{ChatGPT}, we propose distinct system prompts to differentiate between these domains, thereby reducing learning difficulty and accelerating convergence. 
Specifically,  during our proposed three-stage progressive training phase, we use two types of system prompts (“Template A” and “Template B”) that are directly prepended to the image prompt, as shown in \tabref{tab:system_prompt}.
For the unified multi-image generation, we introduce an additional fine-tuning phase (see \secref{training_setup} for details) with the system prompt “Template C”. 

\noindent \textbf{Auxiliary Loss.} 
When training our model on high-resolution images, the model exhibits significant improvements in high-frequency details while some degradations in low-frequency structures. 
We introduce an auxiliary loss to address this issue, which computes the flow-matching objective~\cite{lipman2022flow} with latent features downsampled by a factor of 4:
\begin{align}
    \mathcal{L}_{\text{aux}}(\theta) &= \mathbb{E}_{t,x,\epsilon} \left\| v_{\theta}\left( x_t, t \right) - u_t \right\|^2, 
\end{align}
where \(t ~\in [0,1]\) denotes timestep, \( x = \text{AvgPool}_4(z) \) denotes the downsampled latent features using average pooling by a factor of 4, \( \epsilon \sim \mathcal{N}(0, I) \) is random Gaussian noise, \(v_{\theta}(\cdot)\) and \(u_t = x - \epsilon\) represent the predicted vector field and target vector field, respectively.
This approach helps preserve low-frequency features while learning high-frequency details, enabling efficient knowledge integration and allowing direct fine-tuning at 1024 resolution. 

\subsection{Efficient Inference}
To boost the sampling speed as much as possible while maintaining high sampling quality, Lumina-Image 2.0 makes a deeper exploration on inference efficiency. 

\noindent \textbf{CFG-Renormalization (CFG-Renorm).} Classifier-free guidance (CFG)~\cite{ho2022classifier} is known for improving both visual quality and text-image alignment. During inference, at each timestep $t$, the predicted velocity $v_t$ is calculated as $v_t = v_{t_u} + w(v_{t_c} - v_{t_u})$, where $w$ is the CFG scale, $v_{t_c}$ and $v_{t_u}$ represent the conditional and unconditional velocity, respectively. 
However, scaling by a large $w$ may introduce extremely high activations in certain dimensions of $v_t$, and these abnormal values can result in visual artifacts in the final generated samples.
To address this, recent work introduces the CFG-Renorm method~\cite{lin2024stiv} to rescale the magnitude of the modified velocity of $v_t$ using that of the conditional velocity $v_{t_c}$. We find that this technique effectively improves the stability of CFG-guided generation without introducing additional computational costs.

\noindent \textbf{CFG-Truncation (CFG-Trunc).}  Recent research~\cite{yi2024towards} indicates that text information is largely captured in the early generation stages. Therefore, evaluating $v_{t_c}$ beyond the early timesteps may be redundant. 
The CFG-Trunc can be formulated as follows:
\begin{equation}
    v_t = \left\{\begin{matrix}
    v_{t_u} + w(v_{t_c} - v_{t_u}) & t\ge \alpha;
    \\
    v_{t_u} & t < \alpha.
\end{matrix}\right.
\end{equation}
where $\alpha$ denotes a predefined threshold. This modification can achieve over a 20\% acceleration in sampling speed, without visual degradation.

\noindent \textbf{Flow-DPM-Solver (FDPM).} Lumina-Next supports a range of ODE solvers, such as Midpoint and Euler method. 
While these solvers ensure stability, they are relatively slow since they are not designed for flow models, requiring a large number of function evaluations (NFE) for convergence. 
To improve this, we integrate FDPM~\cite{xie2024sana,lu2022dpm}, which modify DPM-Solver++~\cite{lu2022dpm} to flow models, into Lumina-Image 2.0. FDPM achieves convergence in just 14-20 NFEs, providing a faster sampling solution. 
However, we find that FDPM sometimes suffers from poor stability in practice. 

\noindent \textbf{Timestep Embedding Aware Cache (TeaCache).} TeaCache~\cite{liu2024timestep} is designed to selectively cache informative intermediate results during the inference, thereby accelerating diffusion models. 
TeaCache has successfully accelerated various mainstream image and video generation models, including FLUX~\cite{flux2023}, HunyuanVideo~\cite{kong2024hunyuanvideo}, as well as Lumina-Next. 
Building on its success, we integrate TeaCache into Lumina-Image 2.0. However, our experiments show that TeaCache also introduces visual quality degradations when combined with the above techniques.

\noindent \textbf{Discussion.} 
\label{sec:infer_dis}
The above four inference strategies are mutually compatible and can be applied in combination. 
Notably, Lumina-Image 2.0 is the first to demonstrate that CFG-Renorm and CFG-Trunc provide complementary benefits when applied together. 
CFG-Renorm addresses the issue of over-saturation and visual artifacts when the CFG scale is large, while CFG-Trunc further alleviates this problem by eliminating redundant CFG calculations and achieving acceleration at the same time. 
The flexibility of the CFG scale can be significantly extended to a wider range by combining these techniques.
FDPM and TeaCache can also be integrated into the pipeline, but both of them present certain challenges. 
FDPM lacks sufficient stability and frequently produces suboptimal samples while TeaCache results in blurriness in the sampled images. 
For further details, refer to \secref{sec:infer_ablation}.

\section{Experiments}

\subsection{Implement Details}

\noindent \textbf{Training Dataset.}
\label{training_dataset}
Following the methods in \cite{xie2024show,sun2024autoregressive,wang2024emu3,esser2024scaling,chen2023pixart,xiao2024omnigen,kirstain2023pick}, we constructed a dataset combining both real and synthetic data, and performed data filtering based on the techniques outlined in \cite{chen2023pixart,li2024hunyuan,kong2024hunyuanvideo}, resulting in total 110M samples. 
This dataset is reorganized into three training phases, with 100M, 10M, and 1M samples for each training phase. As the dataset size decreased, the quality of the data progressively improved.


\noindent \textbf{Architecture and Training Setups.}
\label{training_setup}
The architecture configurations of our Unified Next-DiT model, along with a comparison to Lumina-Next~\cite{zhuo2024lumina}, are summarized in  \tabref{tab:model_hyperparams}.
We employed 32 A100 GPUs across all three stages to optimize our Unified Next-DiT. The corresponding training configurations are detailed in \tabref{tab:train_config}.
In addition, for multi-image generation task, we introduce an extra fine-tuning phase, where we consolidate different visual tasks into image grids and generate captions for these concatenated grids to form image-pair pairs. Besides, for the UniCap model, we finetune the Qwen2-VL-7B~\cite{qwen2vl} based on the constructed caption dataset with multi-domain visual data and diverse textual descriptions.

\begin{table*}[!h]
\centering
\setlength{\tabcolsep}{2.3pt} 
\scriptsize  
\caption{Comparison of configuration between Lumina-Next and Lumina-Image 2.0.}
\resizebox{1\linewidth}{!}{ 
\begin{tabular}{l c c c c c c c c}
\toprule
\textbf{Model} & \textbf{Params} & \textbf{Patch Size} & \textbf{Dimension} & \textbf{Heads} & \textbf{KV Heads} & \textbf{Layers} & \textbf{RMSNorm $\epsilon$ \cite{zhang2019root}} & \textbf{Pos. Emb.} \\ 
\midrule
Lumina-Next & 1.7B & 2 & 2304 & 16 & 8 & 24 & $1e^{-5}$ & 2D-RoPE \\
\textbf{Lumina-Image 2.0} & 2.6B & 2 & 2304 & 24 & 8 & 26 & $1e^{-5}$ & M-RoPE \\
\bottomrule
\end{tabular}
}
\label{tab:model_hyperparams}
\end{table*}
 
\begin{table*}[!h]
\centering
\renewcommand{\arraystretch}{1.2}  
\caption{Training configuration across different stages.}
\resizebox{\textwidth}{!}{%
\begin{tabular}{l c c c c c c c}
\toprule
\textbf{Stage} & \textbf{Image Resolution} & \textbf{\#Images} & \textbf{Training Steps (K)} & \textbf{Batch Size} & \textbf{Learning Rate} & \textbf{GPU Days (A100)} & \textbf{Optimizer} \\ 
\midrule
Low Res. Stage    & 256$\times$256  & 100M  & 144  & 1024  & $2 \times 10^{-4}$  & 191  & \multirow{3}{*}{AdamW~\cite{loshchilov2017decoupled}} \\ 
High Res. Stage   & 1024$\times$1024  & 10M   & 40  & 512 & $2 \times 10^{-4}$  & 176 &  \\ 
HQ Tuning Stage   & 1024$\times$1024  & 1M    & 15   & 512 & $2 \times 10^{-4}$  & 224 &  \\ 
\bottomrule
\end{tabular}%
}
\label{tab:train_config}
\end{table*}


\subsection{Quantitative Performance}

\begin{table*}[!t]
\centering
\renewcommand{\arraystretch}{1.1}  
\vspace{-0.2cm}
\caption{Performance comparison across different models on GenEval \cite{ghosh2024geneval}, DPG \cite{hu2024ella}, and T2I-CompBench \cite{huang2023t2i} benchmarks. "$\downarrow$" or "$\uparrow$" indicate lower or higher values are better. \textbf{Bold} indicates the best performance, while \underline{underlining} denotes the second-best performance.}
\resizebox{\linewidth}{!}{%
\begin{tabular}{l c cccc cccc ccc}
\toprule
\multirow{2}{*}{\textbf{Methods}} & \multirow{2}{*}{\textbf{\# Params}} & \multicolumn{4}{c}{\textbf{GenEval} $\uparrow$} & \multicolumn{4}{c}{\textbf{DPG} $\uparrow$} & \multicolumn{3}{c}{\textbf{T2I-CompBench} $\uparrow$} \\ 
\cmidrule(lr){3-6} \cmidrule(lr){7-10} \cmidrule(lr){11-13}
& & Two Obj. & Counting & Color Attri. & \textbf{Overall} & Entity & Relation & Attribute & \textbf{Overall} & Color & Shape & Texture \\ 
\midrule
\multicolumn{13}{l}{\textbf{AutoRegressive Models}} \\ \hline
LlamaGen~\cite{sun2024autoregressive}    & 0.8B & 0.34  & 0.21  & 0.04  & 0.32  & -      & -      & -      & 65.16  & -      & -      & -      \\ 
Chameleon~\cite{team2024chameleon}   & 7B   & -     & -     & -     & 0.39  & -      & -      & -      & -      & -      & -      & -      \\ 
HART~\cite{tang2024hart}       & 732M & -     & -     & -     & -     & -      & -      & -      & 80.89  & -      & -      & -      \\ 
Show-o~\cite{xie2024show}     & 1.3B & 0.52  & 0.49  & 0.28  & 0.53  & -      & -      & -      & 67.48  & -      & -      & -      \\ 
Emu3~\cite{wang2024emu3}        & 8.0B & 0.81 & 0.49 & 0.45 & 0.66 & 87.17 & 90.61 & 86.33 & 81.60  & 0.7544 & 0.5706 & 0.7164 \\ 
Infinity~\cite{han2024infinity}  & 2B  & 0.85 & - & 0.57 & 0.73 & - & 90.76 & - & 83.46 & - & - & - \\ 
Janus-Pro-1B~\cite{chen2025januspro}  & 1.5B  & 0.82 & 0.51 & 0.56 &  0.73 & 88.63 & 88.98 & 88.17 &  82.63 & - & - & - \\ 
Janus-Pro-7B~\cite{chen2025januspro} & 7B  & \textbf{0.89} & 0.59 & \textbf{0.66} &  \textbf{0.80} & 88.90 & 89.32 & \underline{89.40} &  84.19 & - & - & - \\ 
\midrule
\multicolumn{13}{l}{\textbf{Diffusion Models}} \\ \hline
LDM~\cite{rombach2022high}        & 1.4B & 0.29  & 0.23  & 0.05  & 0.37  & -      & -      & -      & 63.18  & -      & -      & -      \\ 
SDv1.5~\cite{rombach2022high}      & 0.9B & -  & -  & -  & 0.40  & 74.23  & 73.49  & 75.39  & 63.18  &0.3730 &0.3646 &0.4219      \\ 
Lumina-Next~\cite{zhuo2024lumina} & 1.7B  & 0.49  &  0.38  &  0.15  & 0.46  & 83.78      &89.78     & 82.67      & 75.66      &0.5088 &0.3386 &0.4239      \\ 
SDv2.1~\cite{rombach2022high}     & 0.9B & 0.51  & 0.44  & 0.50  & 0.47  & -  & -  & -  & 68.09  &0.5694 &0.4495 &0.4982      \\ 
PixArt-$\alpha$~\cite{chen2023pixart} & 0.6B & 0.50 & 0.44  & 0.07  & 0.48  & 79.32  & 82.57  & 78.60  & 71.11  &0.6886 &0.5582 &0.7044     \\ 
SDXL~\cite{podell2023sdxl}  & 2.6B & 0.74  & 0.39  & 0.23  & 0.55  & 82.43  & 86.76  & 80.91  & 74.65  &0.6369 &0.5408 &0.5637      \\ 
SD3-medium~\cite{esser2024scaling} & 2B   & 0.74  & 0.63  & 0.36     & 0.62  &  91.01      &   80.70      & 88.83      & 84.08  & -      & -      & -      \\ 
JanusFlow~\cite{ma2024janusflow}  & 1.3B  & 0.59 & 0.45 & 0.42 &  0.63 & 87.31 & 89.79 & 87.39 &  80.09 & - & - & - \\ 
Sana-0.6B~\cite{xie2024sana} & 0.6B  & 0.76  &  0.64  & 0.39  & 0.64  & 89.50      &  90.10      & 89.30      & 83.60      & -      & -      & -      \\ 
Sana-1.6B~\cite{xie2024sana} & 1.6B  & 0.77  &  0.62  &  0.47  & 0.66  & \underline{91.50}      &\underline{91.90}      & 88.90      & \underline{84.80}      & -      & -      & -      \\ 
DALL-E3~\cite{betker2023improving}   & -  & 0.87    & 0.47     & 0.45  & 0.67 &89.61 & 90.58 &88.39 & 83.50 &\underline{0.8110} &\textbf{0.6750} &\textbf{0.8070} \\ 
OmniGen~\cite{xiao2024omnigen}  & 3.8B  & 0.86 & 0.64 & 0.55 &  0.70 & - & - & - &  - & - & - & - \\ 
Sana-1.5~\cite{xie2025sana}  & 4.8B & 0.85  & \textbf{0.77}  & 0.54  & 0.72  & -  & -  & -  & 85.00  &- &- &-      \\ 
\rowcolor{gray!10} 
\textbf{Lumina-Image 2.0} & 2.6B  & \underline{0.87} & \underline{0.67} & \underline{0.62} &\underline{0.73} & \textbf{91.97} & \textbf{94.85} & \textbf{90.20} & \textbf{87.20} &\textbf{0.8211} &\underline{0.6028} &\underline{0.7417} \\ 
\bottomrule
\end{tabular}%
    }
\vspace{-0.2cm}
\label{tab:performance_comparison}
\end{table*}

\noindent \textbf{Main Results.} We evaluate our model on three benchmarks: GenEval~\cite{ghosh2024geneval}, DPG~\cite{hu2024ella}, and T2I-CompBench~\cite{huang2023t2i}. As shown in \tabref{tab:performance_comparison}, Our model demonstrates strong performance across various metrics on the GenEval benchmark. In the Two Object, Counting, Color Attribute, and Overall metrics, we achieve the second-best performance compared to autoregressive and diffusion models.
On the DPG benchmark, Lumina-Image 2.0 outperforms all compared models across three sub-metrics (Entity, Relation, and Attribute) as well as the Overall metric. Similarly, on T2I-CompBench, our model achieves the best results in both Color and Shape metrics.
The significant advantage we achieve on the DPG benchmark is attributed to the detailed and accurate captions curated by our carefully designed captioning system. Our UniCap generates extremely long and detailed descriptions, which align with the characteristics of prompts contained in DPG, resulting in the strong performance across various metrics, especially in the Relation score.

\begin{table}[!h]
\centering
\begin{minipage}{0.53\textwidth}
\centering
\setlength{\tabcolsep}{10pt} 
\renewcommand{\arraystretch}{1.4}  
\caption{Comparison of ELO scores evaluated in text-to-image arena from Artificial Analysis \footref{fn:artificial_analysis} (as of February 23, 2025).}
\resizebox{\linewidth}{!}{%
\begin{tabular}{l cccc}
\toprule
\multirow{2}{*}{\textbf{Methods}} & \multicolumn{4}{c}{\includegraphics[width=0.5cm]{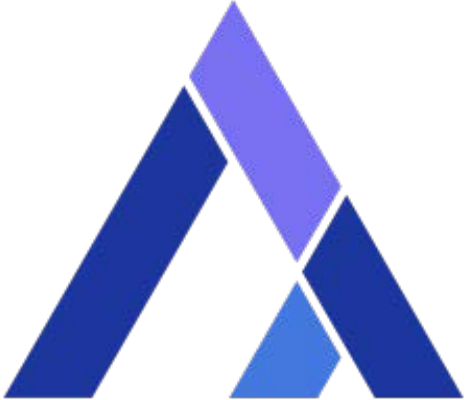} \textbf{Artificial Analysis}} \\
\cmidrule(lr){2-5}
 & Overall & Traditional Art & Fantasy \& Mythical & Anime \\
\midrule
FLUX1.1 [pro] \cite{flux2023} & 1122 & 1075 & 1111 & 1127 \\
FLUX1 [pro] \cite{flux2023} & 1107 & 983 & 1081 & 1051 \\
\rowcolor{gray!19} 
\textbf{Lumina-Image 2.0} & 982 & 1015 & 1051 & 1037 \\
DALLE 3 \cite{betker2023improving} & 970 & 1008 & 1026 & 977 \\
SD3 Medium \cite{esser2024scaling} & 945 & 990 & 1026 & 929 \\
Janus Pro \cite{chen2025januspro} & 748 & 828 & 784 & 766 \\
\bottomrule
\end{tabular}%
}
\label{tab:model_comparison_artificial}
\end{minipage}%
\hfill
\begin{minipage}{0.43\textwidth}
\centering
\setlength{\tabcolsep}{10pt} 
\renewcommand{\arraystretch}{1.4}  
\caption{Comparison of ELO scores evaluated in text-to-image arena from Rapidata \footref{fn:rapidata} (as of February 23, 2025).}
\resizebox{\linewidth}{!}{%
\begin{tabular}{l ccc}
\toprule
\multirow{2}{*}{\textbf{Methods}} & \multicolumn{3}{c}{\includegraphics[width=0.3cm]{Figure/rapidata_log.pdf} \textbf{Rapidata}} \\
\cmidrule(lr){2-4}
 & Overall & Alignment & Coherence \\
\midrule
FLUX1.1 [pro] \cite{flux2023} & 1040 & 1036 & 1023 \\
Imagen 3 \cite{baldridge2024imagen} & 1018 & 1003 & 1032 \\
\rowcolor{gray!19} 
\textbf{Lumina-Image 2.0} & 969 & 1031 & 986 \\
DALLE 3 \cite{betker2023improving} & 952 & 1022 & 958 \\
SD3 Medium \cite{esser2024scaling} & 952 & 1022 & 984 \\
Janus Pro \cite{chen2025januspro} & 734 & 932 & 947 \\
\bottomrule
\end{tabular}%
}
\label{tab:model_comparison_rapidata}
\end{minipage}
\end{table}

\noindent \textbf{Compaison with ELO Scores.}
To better evaluate our model, we present evaluation results from three text-to-image arenas, with all ELO scores~\cite{elo1978rating} based on ratings from human annotators. 
(1) We first perform tests on Artificial Analysis\footnote{\url{https://artificialanalysis.ai/text-to-image/arena?tab=Leaderboard} \label{fn:artificial_analysis}}. 
As shown in~\tabref{tab:model_comparison_artificial}, Lumina-Image 2.0 achieves mid-tier results, outperforming almost all open-source models (e.g., SD3~\cite{esser2024scaling} and Janus Pro~\cite{chen2025januspro}) and several closed-source systems (e.g., DALL·E 3~\cite{betker2023improving}), but still lags behind the top closed-source models, such as FLUX Pro~\cite{flux2023}. 
(2) To analyze the alignment and coherence abilities of our model, we also provide rankings from Rapidata\footnote{\url{https://www.rapidata.ai/leaderboard/image-models} \label{fn:rapidata}}.
As shown in~\tabref{tab:model_comparison_rapidata}, our model achieves a comparable ranking. 
In particular, Lumina-Image 2.0 ranks second only to FLUX Pro in terms of prompt alignment, exceeding many other closed-source models such as Imagen~3~\cite{baldridge2024imagen}. 
This further validates the effectiveness of our proposed Unified Next-DiT architecture and UniCap annotation system. 
Moreover, although Janus Pro\cite{chen2025januspro} achieves state-of-the-art results on academic benchmarks, its scores were considerably lower than those of Lumina-Image 2.0 and FLUX Pro on user-driven leaderboards.
This discrepancy highlights the inherent bias and limitations in current academic benchmarks. 
(3) Finally, results from AGI-Eval \footnote{\url{https://ai-ceping.com/} \label{fn:ce_ping}} in \tabref{tab:model_comparison_meituan} demonstrate that Lumina-Image 2.0 significantly outperforms the previous Lumina-Next~\cite{zhuo2024lumina} as well as all other Chinese open-source models~\cite{kolors2024,li2024hunyuan}. 

In summary, we hope that this comprehensive evaluation and comparative analysis will provide the community with a clearer understanding of Lumina-Image 2.0's capabilities and constraints, thereby guiding future improvements. 
We also believe that developing better human-aligned evaluation benchmarks is essential to accurately assess current models and advance generative modeling progress.

\begin{table}[H]
\centering
\setlength{\tabcolsep}{10pt} 
\renewcommand{\arraystretch}{1.4}  
\caption{Comparison of ELO scores evaluated in text-to-image arena from AGI-Eval \footref{fn:ce_ping} (as of February 23, 2025).}
\resizebox{0.9\linewidth}{!}{%
\begin{tabular}{l c c c c c c c}
\toprule
\textbf{Model} & FLUX1.1 [pro] \cite{flux2023} &FLUX.1 [dev] \cite{flux2023} & \textbf{Lumina-Image 2.0}  & Kolors \cite{kolors2024} &HunyuanDiT \cite{li2024hunyuan} &Lumina-Next \cite{zhuo2024lumina} \\
\midrule
\textbf{Score} &0.4859 &0.4712 & 0.4545 &0.3924 & 0.3920 & 0.3229 \\
\bottomrule
\end{tabular}%
}
\label{tab:model_comparison_meituan}
\end{table}


\subsection{Qualitative Performance}

\begin{figure}[!t]
    \centering
    \includegraphics[scale=0.48]{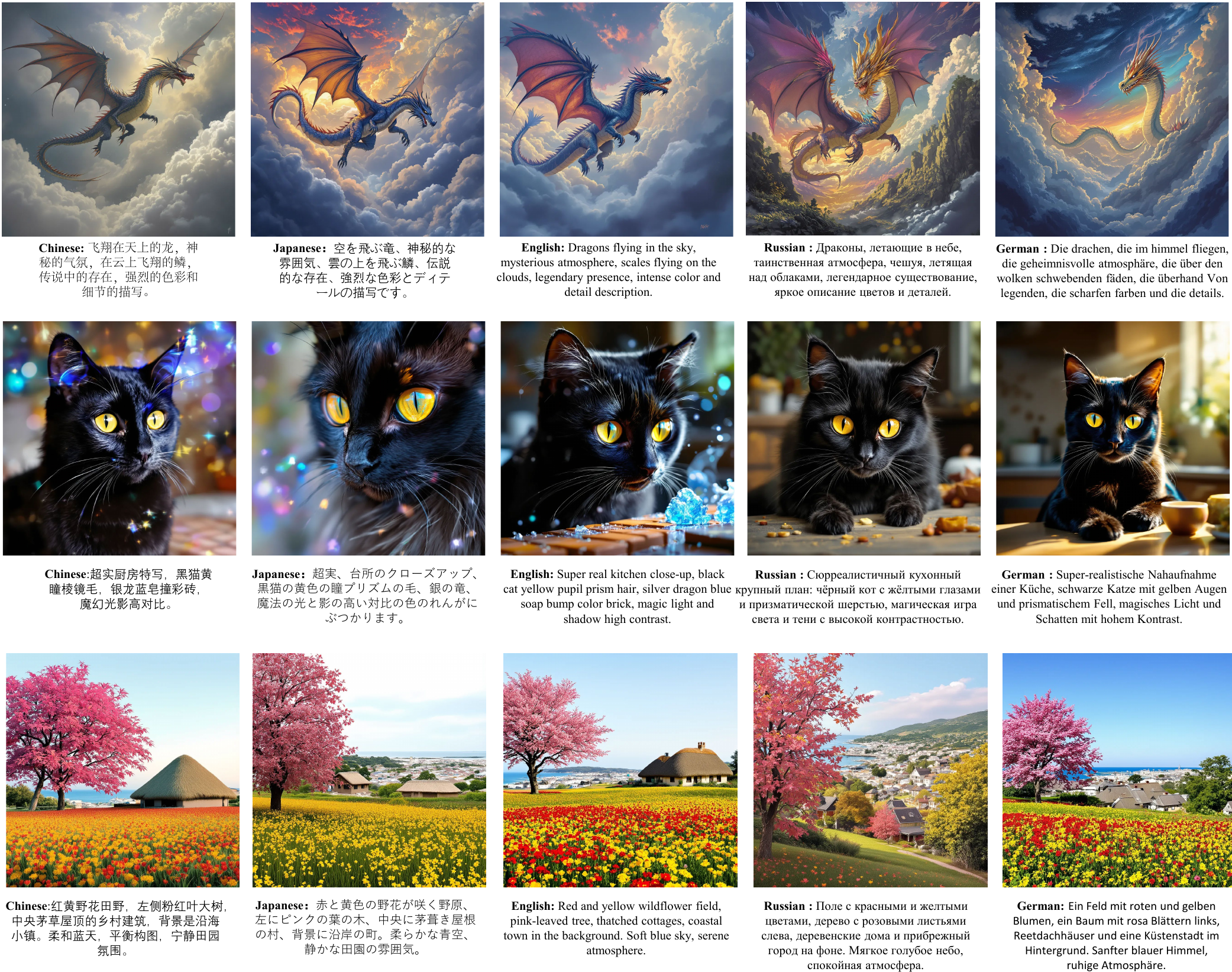}
    \caption{Visualization results of multilingual text-to-image generation by our Lumina-Image 2.0, covering five languages: Chinese, Japanese, English, Russian, and German.}
    \label{fig:multilingual}
\end{figure}

\textbf{Multi-lingual Generation.} 
Compared to previous T2I models \cite{chen2023pixart,saharia2022photorealistic} that use CLIP \cite{radford2021learning} and T5 \cite{2020t5} as text encoders, we employ Gemma2-2b~\cite{team2024gemma} as the text encoder, enabling our model to understand multiple languages. 
It naturally exhibits zero-shot capability in languages such as German, Japanese, and Russian. 
As shown in \figref{fig:multilingual}, we present the generation results in five different languages.

\begin{figure}[!t]
    \centering
    \includegraphics[scale=0.157]{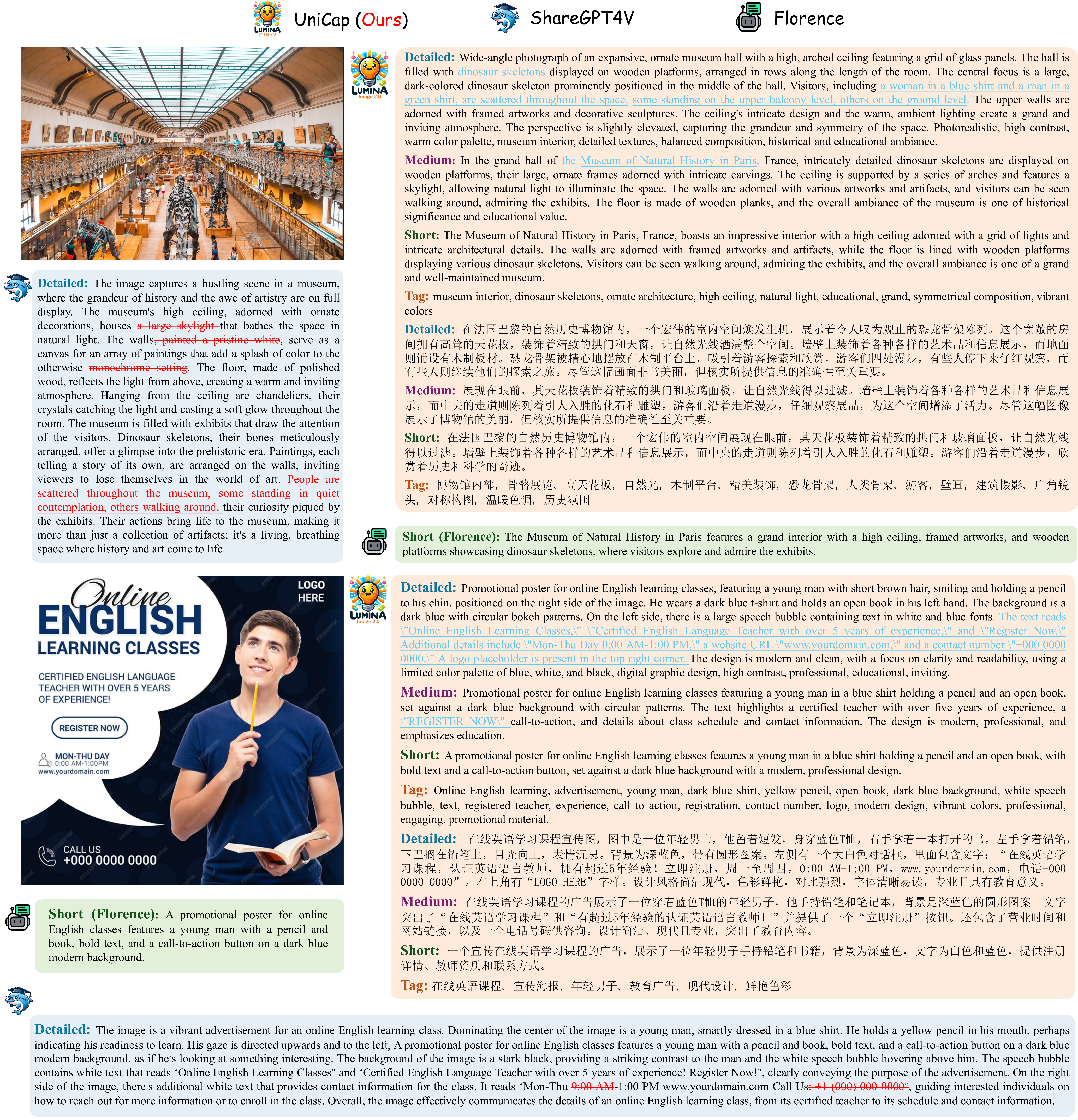}
    \caption{Comparison with ShareGPT4V \cite{chen2024sharegpt4v} and Florence \cite{xiao2024florence} in complex scenes and dense text for caption generation. The \textcolor{wblue}{\underline{blue underline}} correspond to areas with more detailed and accurate descriptions, while \textcolor{red}{\underline{red underline}} and \textcolor{red}{\sout{red strikethrough}} represent the incorrect and insufficient descriptions respectively. }
    \label{fig:captioner_1}
\end{figure}

\begin{figure}[!th]
    \centering
    \includegraphics[scale=0.157]{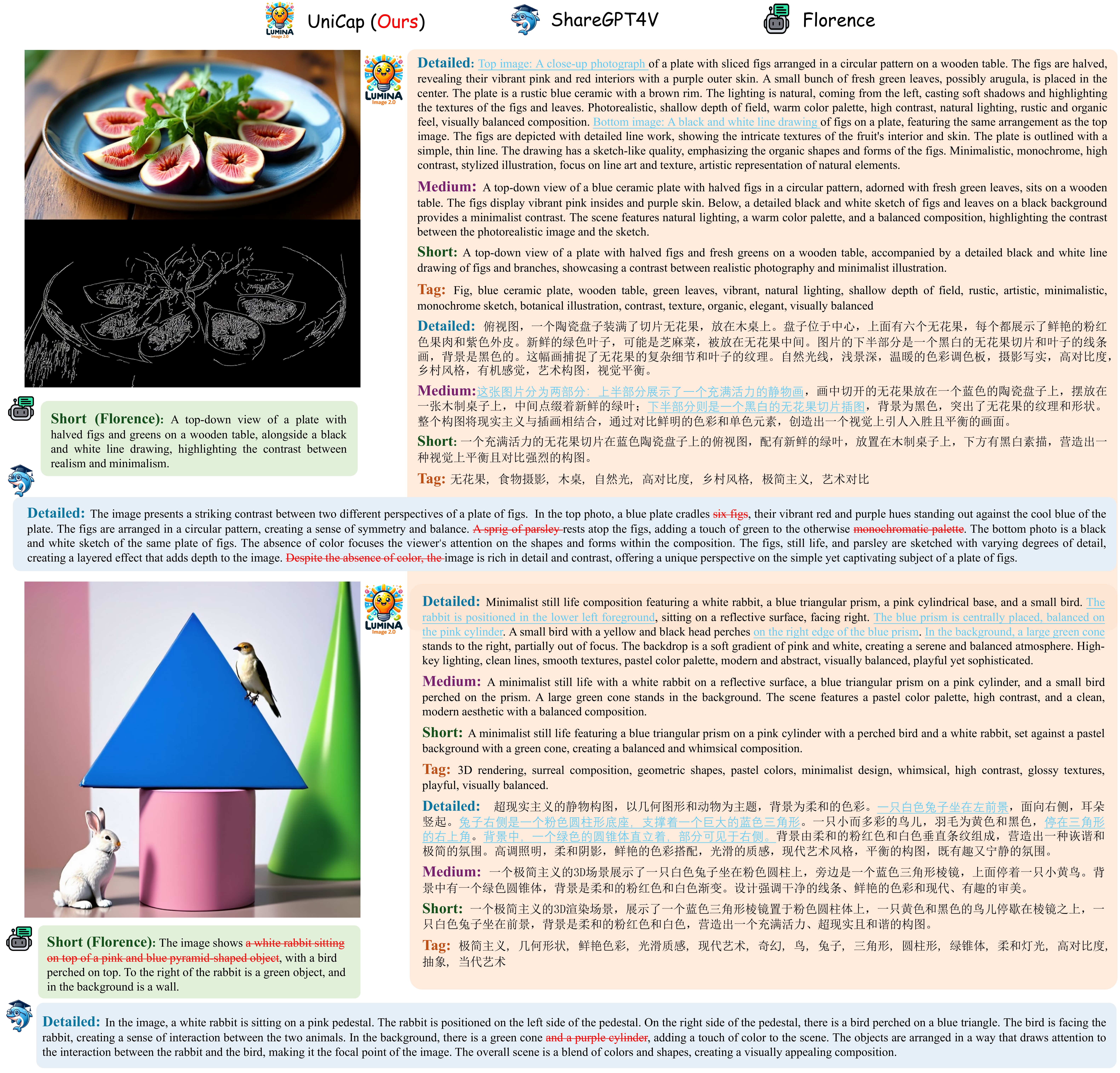}
    \caption{Comparison with ShareGPT4V \cite{chen2024sharegpt4v} and Florence \cite{xiao2024florence} in visual understanding and spatial relationships. The \textcolor{wblue}{\underline{blue underline}} correspond to areas with more detailed and accurate descriptions, while \textcolor{red}{\underline{red underline}} and \textcolor{red}{\sout{red strikethrough}} represent the incorrect and insufficient descriptions respectively.}
    \label{fig:captioner_2}
\end{figure}

\textbf{Captioning Everything With UniCap.}
We compare our proposed UniCap with existing captioners, such as ShareGPT4V \cite{chen2024sharegpt4v} and Florence \cite{xiao2024florence}, from four dimensions: complex scenes, dense text, visual understanding, and spatial relationships. UniCap supports multilingual annotations, including both Chinese and English, and can generate captions of varying lengths to accommodate diverse user needs. As shown in \figref{fig:captioner_1} and \figref{fig:captioner_2}, UniCap delivers highly detailed and accurate descriptions, significantly outperforming the other two methods.

\begin{figure}[!th]
    \centering
    \includegraphics[scale=0.146]{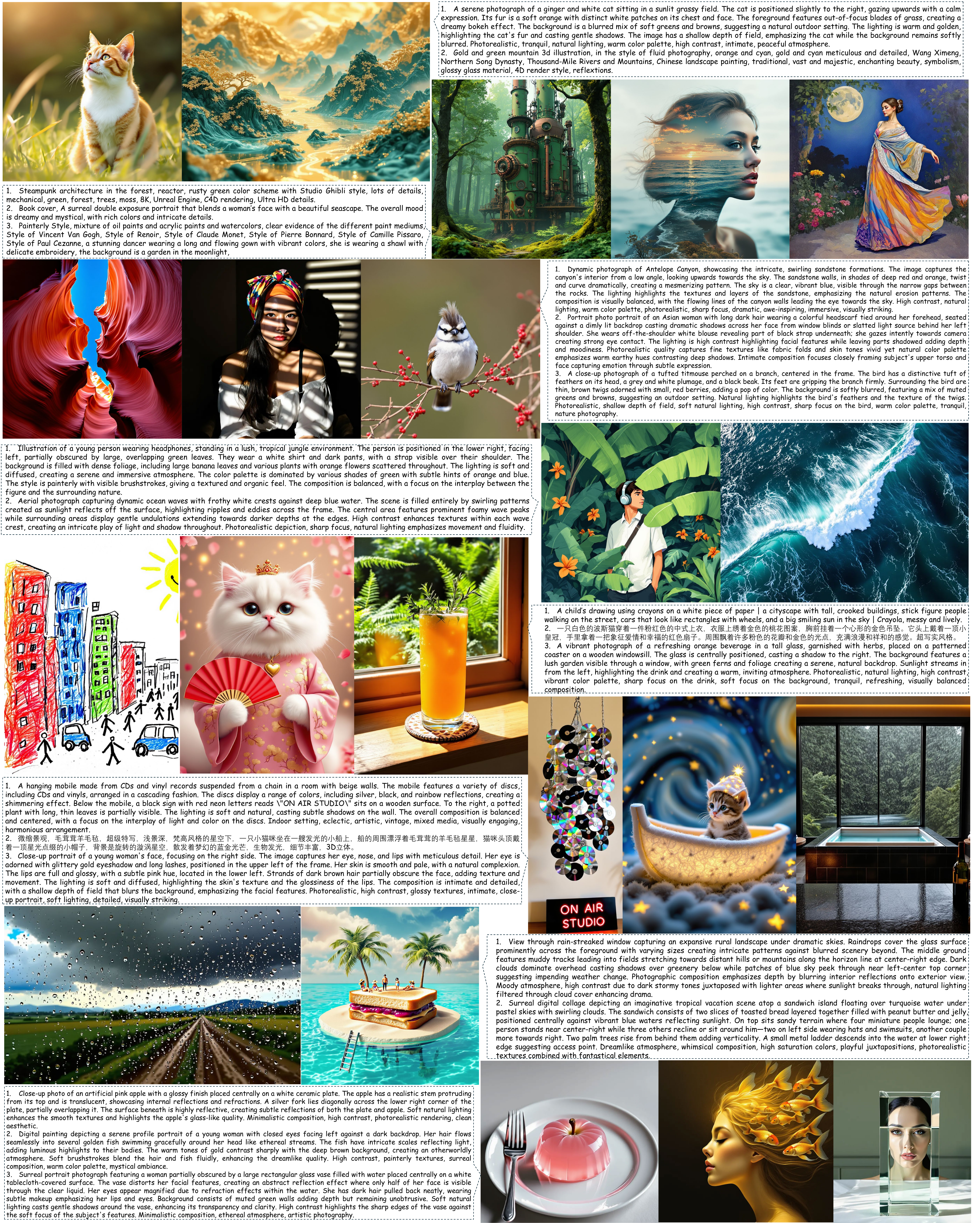}
    \caption{High-quality image generation examples from Lumina-Image 2.0, showcasing its precise prompt-following ability and its capability to generate highly aesthetic and realistic images across different resolutions.}
    \label{fig:demos_2}
\end{figure}

\textbf{High-quality Image Generation.}
In~\figref{fig:demos_2}, we present additional generation results of Lumina-Image 2.0. These results illustrate that our model is capable of producing high-quality images in various resolutions that are visually realistic, highly aesthetic, and creatively expressive. Furthermore, extensive experiments with both Chinese and English prompts of different lengths demonstrate robust text-image alignment.

\subsection{Ablation Study}

\begin{figure}[!ht]
    \centering
    \includegraphics[scale=0.4]{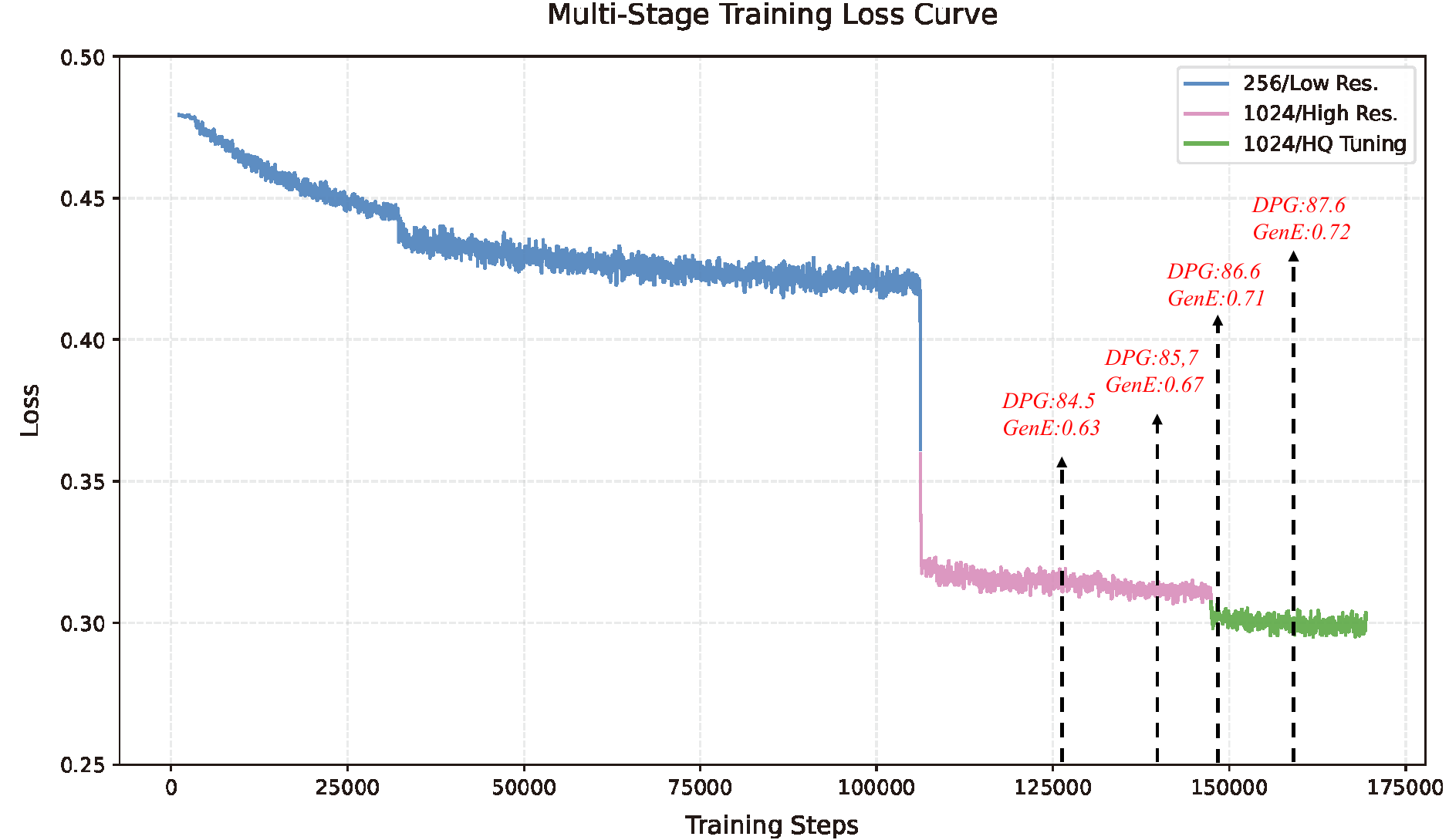}
    \caption{Loss curves for the three training stages, showing a steady performance increase in the DPG~\cite{hu2024ella} and GenEval~\cite{ghosh2024geneval} benchmark.}
    \label{fig:loss_stage1}
\end{figure}

\noindent \textbf{Ablation Study on Multi-Stage Training Strategy.} During our three-stage progressive training, the model’s performance steadily improves as dataset size decreased and quality increased.
As shown in \tabref{tab:ab_multi-stage} and \figref{fig:loss_stage1}, the performances continue to advance from the second to the third stage, evidenced by both quantitative improvements and loss curve trends. 
In the high-quality tuning stage, the model achieves substantial improvements within just 1K steps, e.g. from $85.7$ to $86.6$ on the DPG benchmark and from $0.67$ to $0.71$ on the GenEval. 
However, as high-quality tuning progressed, performance fluctuations are observed. 
Notably, at 11K steps in the third stage, the model continues improving on DPG benchmark, whereas the performance declines slightly on GenEval.

\begin{table}[!h]
\centering
\vspace{-0.1cm}
\caption{Performance Comparison Across Stages on DPG~\cite{hu2024ella} and GenEval~\cite{ghosh2024geneval} Benchmarks.}
\vspace{0.1cm}
\renewcommand{\arraystretch}{1.1}  
\setlength{\tabcolsep}{24pt} 
\resizebox{0.7\linewidth}{!}{%
\begin{tabular}{c c c c}
\toprule
\textbf{Stage} & \textbf{Steps (K)} & \textbf{DPG} & \textbf{GenEval} \\ 
\midrule 
Low Res. Stage & 15  & 84.5  & 0.63 \\

High Res. Stage & 38  & 85.7  & 0.67 \\ 
HQ Tuning Stage & 1  & 86.6  & 0.71 \\ 
HQ Tuning Stage & 5  & 87.2  & 0.73 \\ 
HQ Tuning Stage & 11  & 87.6  & 0.72 \\ 
\bottomrule
\end{tabular}%
}
\vspace{-0.2cm}
\label{tab:ab_multi-stage}
\end{table}

\begin{figure}[!t]
    \centering
   \includegraphics[width=0.8\linewidth]{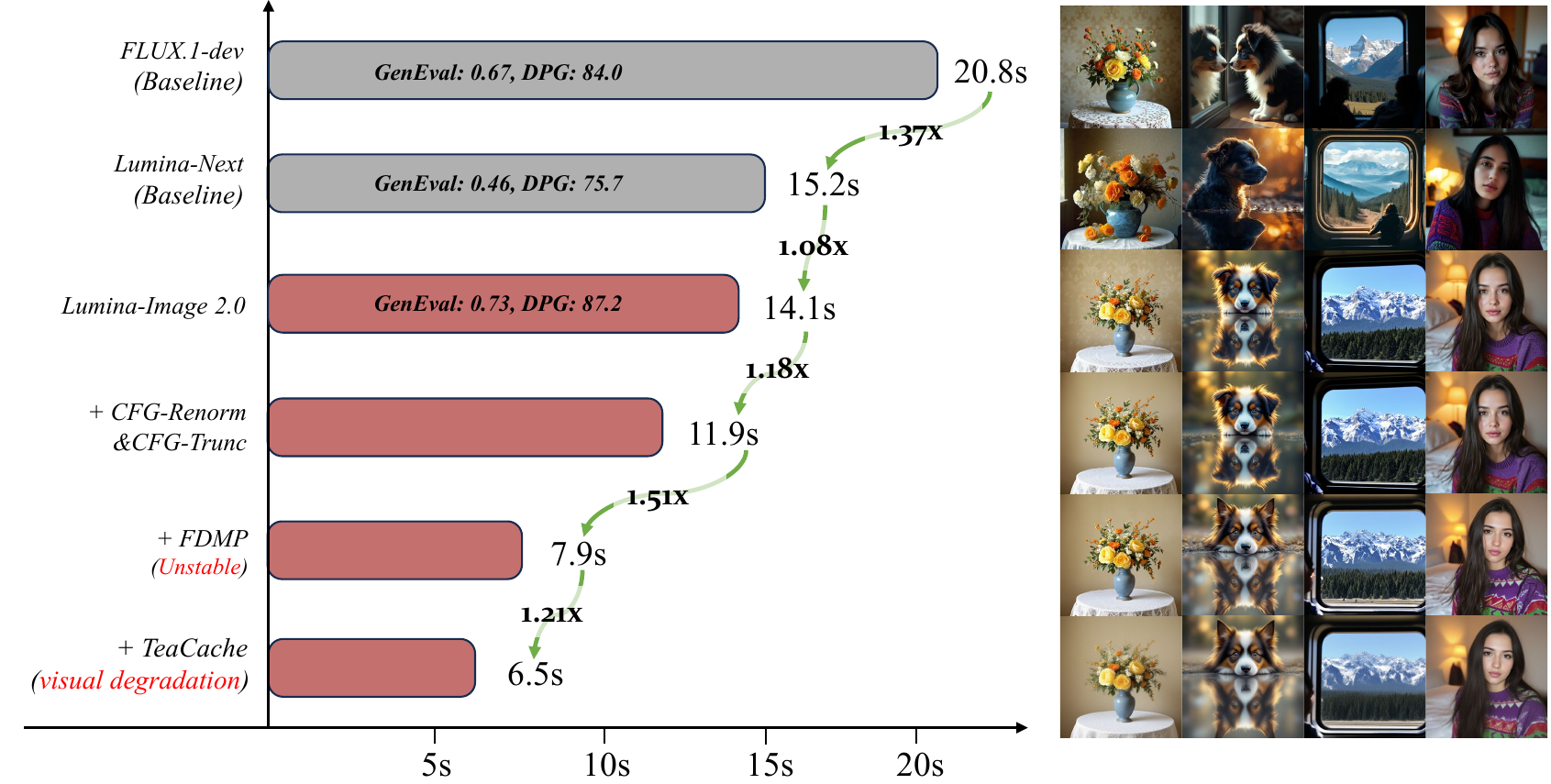}
    \caption{Ablation study on efficient inference strategy. The performances are measured on a single A100 GPU with batch size 1.}
    \label{fig:infer_ab}
\end{figure}

\noindent \textbf{Ablation Study on Efficient Inference Strategy.} 
\label{sec:infer_ablation}
In \figref{fig:infer_ab}, we evaluate the inference efficiency of Lumina-Image 2.0 under a 1024-resolution setting using multiple inference strategies, including CFG-Renorm, CFG-Trunc, FDPM, and TeaCache.  
First, our results show that the proposed CFG-Renorm and CFG-Trunc fusion method (\secref{sec:infer_dis}) not only saves sampling time but also has a negligible impact on the quality of the sampled results.  
Second, integrating FDPM into our model can effectively reduce inference time. However, empirical evaluations indicate that FDPM suffers from poor stability, negatively affecting sample quality during the generation process.  
Third, while incorporating TeaCache further improves sampling speed, it significantly degrades image quality, often leading to blurriness.  
As a result, in practical applications, we adopt Lumina-Image 2.0 with CFG-Renorm and CFG-Trunc as the final solution to balance efficiency and quality.




\section{Limitation}

\begin{figure}[!t]
    \centering
    \includegraphics[scale=0.36]{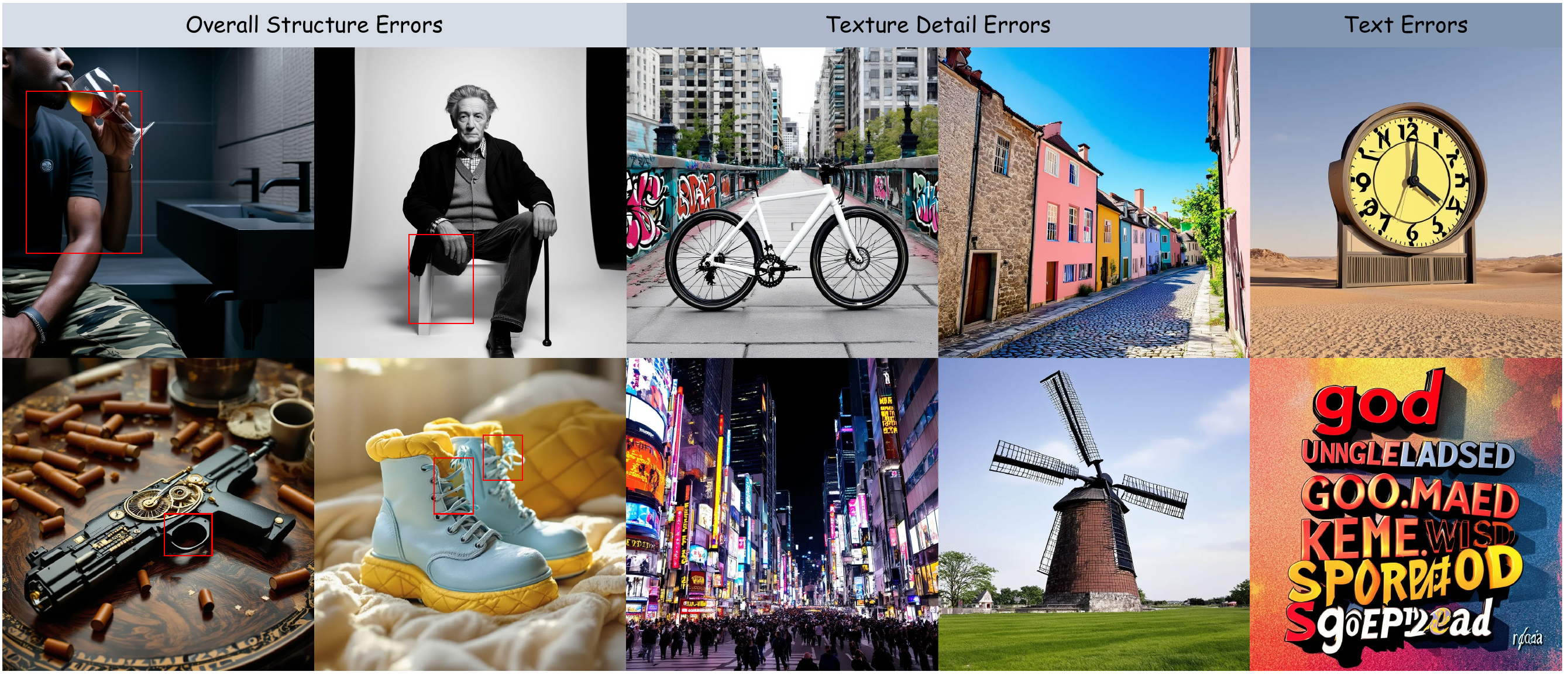}
    \caption{Generation defects of Lumina-Image 2.0, categorized into overall structural errors, texture detail errors, and text errors.}
    \label{fig:bad_case}
\end{figure}

Although we have followed previous works~\cite{xie2024sana,chen2023pixart,han2024infinity,chen2025januspro,xiao2024omnigen} to evaluate our method on benchmarks such as GenEval~\cite{ghosh2024geneval} and T2ICompBench~\cite{huang2023t2i}, achieving comparable performance with state-of-the-art models, we argue that these academic benchmarks are not comprehensive and may sometimes fail to accurately assess image quality in alignment with human perception. 
To illustrate this point, \figref{fig:bad_case} highlights several limitations of Lumina-Image 2.0. 
First, for complex and diverse structures (e.g., human bodies) and for rare concepts in the training data (e.g., handguns), our model struggles to consistently generate correct results. 
Second, when handling images with intricate textures, such as densely crowded scenes or tire spokes, our model frequently generates disordered details. 
Finally, our model still needs substantial improvements in accurately rendering long and complex text.


\vspace{-0.15cm}
\section{Conclusion}
This paper introduces Lumina-Image 2.0, a unified and efficient text-to-image generative framework that achieves strong performance in both image quality and prompt alignment. 
Specifically, a Unified Next-DiT model is developed to generate high-quality images through the seamless integration of textual and visual information. 
A Unified Captioner (UniCap) is proposed to produce detailed and accurate textual descriptions for constructing high-quality image-text training pairs. 
In addition, a set of efficient training and inference strategies is developed to further optimize performance while reducing computational costs. 
Lumina-Image 2.0 achieves promising performance on public benchmarks, and provides a transparent, reproducible text-to-image generative framework. 
We hope that our model will contribute to advancing the field of text-to-image generation.

\bibliographystyle{unsrt}  
\bibliography{references}

\begin{thebibliography}{10}

\bibitem{rombach2022high}
Robin Rombach, Andreas Blattmann, Dominik Lorenz, Patrick Esser, and Bj{\"o}rn Ommer.
\newblock High-resolution image synthesis with latent diffusion models.
\newblock In {\em Proceedings of the IEEE Conference on Computer Vision and Pattern Recognition (CVPR)}, pages 10684--10695, 2022.

\bibitem{betker2023improving}
James Betker, Gabriel Goh, Li~Jing, Tim Brooks, Jianfeng Wang, Linjie Li, Long Ouyang, Juntang Zhuang, Joyce Lee, Yufei Guo, et~al.
\newblock Improving image generation with better captions.
\newblock {\em Computer Science. https://cdn. openai. com/papers/dall-e-3. pdf}, 2(3):8, 2023.

\bibitem{xie2024show}
Jinheng Xie, Weijia Mao, Zechen Bai, David~Junhao Zhang, Weihao Wang, Kevin~Qinghong Lin, Yuchao Gu, Zhijie Chen, Zhenheng Yang, and Mike~Zheng Shou.
\newblock Show-o: One single transformer to unify multimodal understanding and generation.
\newblock {\em arXiv preprint arXiv:2408.12528}, 2024.

\bibitem{sun2024autoregressive}
Peize Sun, Yi~Jiang, Shoufa Chen, Shilong Zhang, Bingyue Peng, Ping Luo, and Zehuan Yuan.
\newblock Autoregressive model beats diffusion: Llama for scalable image generation.
\newblock {\em arXiv preprint arXiv:2406.06525}, 2024.

\bibitem{team2024chameleon}
Chameleon Team.
\newblock Chameleon: Mixed-modal early-fusion foundation models.
\newblock {\em arXiv preprint arXiv:2405.09818}, 2024.

\bibitem{tang2024hart}
Haotian Tang, Yecheng Wu, Shang Yang, Enze Xie, Junsong Chen, Junyu Chen, Zhuoyang Zhang, Han Cai, Yao Lu, and Song Han.
\newblock Hart: Efficient visual generation with hybrid autoregressive transformer.
\newblock {\em arXiv preprint arXiv:2410.10812}, 2024.

\bibitem{wang2024emu3}
Xinlong Wang, Xiaosong Zhang, Zhengxiong Luo, Quan Sun, Yufeng Cui, Jinsheng Wang, Fan Zhang, Yueze Wang, Zhen Li, Qiying Yu, et~al.
\newblock Emu3: Next-token prediction is all you need.
\newblock {\em arXiv preprint arXiv:2409.18869}, 2024.

\bibitem{xie2024sana}
Enze Xie, Junsong Chen, Junyu Chen, Han Cai, Haotian Tang, Yujun Lin, Zhekai Zhang, Muyang Li, Ligeng Zhu, Yao Lu, et~al.
\newblock Sana: Efficient high-resolution image synthesis with linear diffusion transformers.
\newblock {\em Proceedings of the International Conference on Learning Representations (ICLR)}, 2025.

\bibitem{flux2023}
Black~Forest Labs.
\newblock Flux.
\newblock \url{https://github.com/black-forest-labs/flux}, 2023.

\bibitem{chen2025januspro}
Xiaokang Chen, Zhiyu Wu, Xingchao Liu, Zizheng Pan, Wen Liu, Zhenda Xie, Xingkai Yu, and Chong Ruan.
\newblock Janus-pro: Unified multimodal understanding and generation with data and model scaling, 2025.

\bibitem{zhang2023adding}
Lvmin Zhang, Anyi Rao, and Maneesh Agrawala.
\newblock Adding conditional control to text-to-image diffusion models.
\newblock In {\em Proceedings of the IEEE International Conference on Computer Vision (ICCV)}, pages 3836--3847, 2023.

\bibitem{li2024photomaker}
Zhen Li, Mingdeng Cao, Xintao Wang, Zhongang Qi, Ming-Ming Cheng, and Ying Shan.
\newblock Photomaker: Customizing realistic human photos via stacked id embedding.
\newblock In {\em Proceedings of the IEEE Conference on Computer Vision and Pattern Recognition (CVPR)}, pages 8640--8650, 2024.

\bibitem{kawar2023imagic}
Bahjat Kawar, Shiran Zada, Oran Lang, Omer Tov, Huiwen Chang, Tali Dekel, Inbar Mosseri, and Michal Irani.
\newblock Imagic: Text-based real image editing with diffusion models.
\newblock In {\em Proceedings of the IEEE Conference on Computer Vision and Pattern Recognition (CVPR)}, pages 6007--6017, 2023.

\bibitem{lin2024pixwizard}
Weifeng Lin, Xinyu Wei, Renrui Zhang, Le~Zhuo, Shitian Zhao, Siyuan Huang, Junlin Xie, Yu~Qiao, Peng Gao, and Hongsheng Li.
\newblock Pixwizard: Versatile image-to-image visual assistant with open-language instructions.
\newblock {\em arXiv preprint arXiv:2409.15278}, 2024.

\bibitem{chen2023pixart}
Junsong Chen, Jincheng Yu, Chongjian Ge, Lewei Yao, Enze Xie, Yue Wu, Zhongdao Wang, James Kwok, Ping Luo, Huchuan Lu, et~al.
\newblock Pixart-$\alpha$: Fast training of diffusion transformer for photorealistic text-to-image synthesis.
\newblock {\em Proceedings of the International Conference on Learning Representations (ICLR)}, 2023.

\bibitem{gao2024lumin-t2x}
Peng Gao, Le~Zhuo, Chris Liu, , Ruoyi Du, Xu~Luo, Longtian Qiu, Yuhang Zhang, et~al.
\newblock Lumina-t2x: Transforming text into any modality, resolution, and duration via flow-based large diffusion transformers.
\newblock {\em arXiv preprint arXiv:2405.05945}, 2024.

\bibitem{zhuo2024lumina}
Le~Zhuo, Ruoyi Du, Han Xiao, Yangguang Li, Dongyang Liu, Rongjie Huang, Wenze Liu, Lirui Zhao, Fu-Yun Wang, Zhanyu Ma, et~al.
\newblock Lumina-next: Making lumina-t2x stronger and faster with next-dit.
\newblock {\em arXiv preprint arXiv:2406.18583}, 2024.

\bibitem{ma2025step}
Guoqing Ma, Haoyang Huang, Kun Yan, Liangyu Chen, Nan Duan, Shengming Yin, Changyi Wan, Ranchen Ming, Xiaoniu Song, Xing Chen, et~al.
\newblock Step-video-t2v technical report: The practice, challenges, and future of video foundation model.
\newblock {\em arXiv preprint arXiv:2502.10248}, 2025.

\bibitem{chen2025goku}
Shoufa Chen, Chongjian Ge, Yuqi Zhang, Yida Zhang, Fengda Zhu, Hao Yang, Hongxiang Hao, Hui Wu, Zhichao Lai, Yifei Hu, et~al.
\newblock Goku: Flow based video generative foundation models.
\newblock {\em arXiv preprint arXiv:2502.04896}, 2025.

\bibitem{lidit}
Bingqi Ma, Zhuofan Zong, Guanglu Song, Hongsheng Li, and Yu~Liu.
\newblock Exploring the role of large language models in prompt encoding for diffusion models.
\newblock {\em arXiv preprint arXiv:2406.11831}, 2024.

\bibitem{ye2023ip}
Hu~Ye, Jun Zhang, Sibo Liu, Xiao Han, and Wei Yang.
\newblock Ip-adapter: Text compatible image prompt adapter for text-to-image diffusion models.
\newblock {\em arXiv preprint arXiv:2308.06721}, 2023.

\bibitem{li2024hunyuan}
Zhimin Li, Jianwei Zhang, Qin Lin, Jiangfeng Xiong, Yanxin Long, Xinchi Deng, Yingfang Zhang, Xingchao Liu, Minbin Huang, Zedong Xiao, et~al.
\newblock Hunyuan-dit: A powerful multi-resolution diffusion transformer with fine-grained chinese understanding.
\newblock {\em arXiv preprint arXiv:2405.08748}, 2024.

\bibitem{ChatGPT}
OpenAI.
\newblock Chatgpt: Optimizing language models for dialogue, 2022.

\bibitem{jaech2024openai}
Aaron Jaech, Adam Kalai, Adam Lerer, Adam Richardson, Ahmed El-Kishky, Aiden Low, Alec Helyar, Aleksander Madry, Alex Beutel, Alex Carney, et~al.
\newblock Openai o1 system card.
\newblock {\em arXiv preprint arXiv:2412.16720}, 2024.

\bibitem{dubey2024llama}
Abhimanyu Dubey, Abhinav Jauhri, Abhinav Pandey, Abhishek Kadian, Ahmad Al-Dahle, Aiesha Letman, Akhil Mathur, Alan Schelten, Amy Yang, Angela Fan, et~al.
\newblock The llama 3 herd of models.
\newblock {\em arXiv preprint arXiv:2407.21783}, 2024.

\bibitem{touvron2023llama}
Hugo Touvron, Louis Martin, Kevin Stone, Peter Albert, Amjad Almahairi, Yasmine Babaei, Nikolay Bashlykov, Soumya Batra, Prajjwal Bhargava, Shruti Bhosale, et~al.
\newblock Llama 2: Open foundation and fine-tuned chat models.
\newblock {\em arXiv preprint arXiv:2307.09288}, 2023.

\bibitem{lin2024stiv}
Zongyu Lin, Wei Liu, Chen Chen, Jiasen Lu, Wenze Hu, Tsu-Jui Fu, Jesse Allardice, Zhengfeng Lai, Liangchen Song, Bowen Zhang, et~al.
\newblock Stiv: Scalable text and image conditioned video generation.
\newblock {\em arXiv preprint arXiv:2412.07730}, 2024.

\bibitem{yi2024towards}
Mingyang Yi, Aoxue Li, Yi~Xin, and Zhenguo Li.
\newblock Towards understanding the working mechanism of text-to-image diffusion model.
\newblock {\em Advances in Neural Information Processing Systems (NeurIPS)}, 2024.

\bibitem{liu2024timestep}
Feng Liu, Shiwei Zhang, Xiaofeng Wang, Yujie Wei, Haonan Qiu, Yuzhong Zhao, Yingya Zhang, Qixiang Ye, and Fang Wan.
\newblock Timestep embedding tells: It's time to cache for video diffusion model.
\newblock {\em arXiv preprint arXiv:2411.19108}, 2024.

\bibitem{hu2024ella}
Xiwei Hu, Rui Wang, Yixiao Fang, Bin Fu, Pei Cheng, and Gang Yu.
\newblock Ella: Equip diffusion models with llm for enhanced semantic alignment.
\newblock {\em arXiv preprint arXiv:2403.05135}, 2024.

\bibitem{ghosh2024geneval}
Dhruba Ghosh, Hannaneh Hajishirzi, and Ludwig Schmidt.
\newblock Geneval: An object-focused framework for evaluating text-to-image alignment.
\newblock {\em Advances in Neural Information Processing Systems (NeurIPS)}, 36, 2024.

\bibitem{huang2023t2i}
Kaiyi Huang, Kaiyue Sun, Enze Xie, Zhenguo Li, and Xihui Liu.
\newblock T2i-compbench: A comprehensive benchmark for open-world compositional text-to-image generation.
\newblock {\em Advances in Neural Information Processing Systems (NeurIPS)}, 36:78723--78747, 2023.

\bibitem{podell2023sdxl}
Dustin Podell, Zion English, Kyle Lacey, Andreas Blattmann, Tim Dockhorn, Jonas M{\"u}ller, Joe Penna, and Robin Rombach.
\newblock Sdxl: Improving latent diffusion models for high-resolution image synthesis.
\newblock {\em arXiv preprint arXiv:2307.01952}, 2023.

\bibitem{dit}
William~S Peebles and Saining Xie.
\newblock Scalable diffusion models with transformers.
\newblock In {\em Proceedings of the IEEE International Conference on Computer Vision (ICCV)}, volume 4172, 2022.

\bibitem{chen2024pixart}
Junsong Chen, Chongjian Ge, Enze Xie, Yue Wu, Lewei Yao, Xiaozhe Ren, Zhongdao Wang, Ping Luo, Huchuan Lu, and Zhenguo Li.
\newblock Pixart-$\sigma$: Weak-to-strong training of diffusion transformer for 4k text-to-image generation.
\newblock {\em Proceedings of the European Conference on Computer Vision (ECCV)}, 2024.

\bibitem{esser2024scaling}
Patrick Esser, Sumith Kulal, Andreas Blattmann, Rahim Entezari, Jonas M{\"u}ller, Harry Saini, Yam Levi, Dominik Lorenz, Axel Sauer, Frederic Boesel, et~al.
\newblock Scaling rectified flow transformers for high-resolution image synthesis.
\newblock In {\em Proceedings of the International Conference on Machine Learning (ICML)}, 2024.

\bibitem{xiao2024omnigen}
Shitao Xiao, Yueze Wang, Junjie Zhou, Huaying Yuan, Xingrun Xing, Ruiran Yan, Shuting Wang, Tiejun Huang, and Zheng Liu.
\newblock Omnigen: Unified image generation.
\newblock {\em arXiv preprint arXiv:2409.11340}, 2024.

\bibitem{radford2021learning}
Alec Radford, Jong~Wook Kim, Chris Hallacy, Aditya Ramesh, Gabriel Goh, Sandhini Agarwal, Girish Sastry, Amanda Askell, Pamela Mishkin, Jack Clark, et~al.
\newblock Learning transferable visual models from natural language supervision.
\newblock In {\em International conference on machine learning}, pages 8748--8763. PmLR, 2021.

\bibitem{2020t5}
Colin Raffel, Noam Shazeer, Adam Roberts, Katherine Lee, Sharan Narang, Michael Matena, Yanqi Zhou, Wei Li, and Peter~J. Liu.
\newblock Exploring the limits of transfer learning with a unified text-to-text transformer.
\newblock {\em Journal of Machine Learning Research}, 21(140):1--67, 2020.

\bibitem{gemma}
Gemma Team, Thomas Mesnard, Cassidy Hardin, Robert Dadashi, Surya Bhupatiraju, Shreya Pathak, Laurent Sifre, Morgane Rivi{\`e}re, Mihir~Sanjay Kale, Juliette Love, et~al.
\newblock Gemma: Open models based on gemini research and technology.
\newblock {\em arXiv preprint arXiv:2403.08295}, 2024.

\bibitem{liu2022flow}
Xingchao Liu, Chengyue Gong, and Qiang Liu.
\newblock Flow straight and fast: Learning to generate and transfer data with rectified flow.
\newblock {\em arXiv preprint arXiv:2209.03003}, 2022.

\bibitem{lipman2022flow}
Yaron Lipman, Ricky~TQ Chen, Heli Ben-Hamu, Maximilian Nickel, and Matt Le.
\newblock Flow matching for generative modeling.
\newblock {\em arXiv preprint arXiv:2210.02747}, 2022.

\bibitem{liu2024lumina}
Dongyang Liu, Shitian Zhao, Le~Zhuo, Weifeng Lin, Yu~Qiao, Hongsheng Li, and Peng Gao.
\newblock Lumina-mgpt: Illuminate flexible photorealistic text-to-image generation with multimodal generative pretraining.
\newblock {\em arXiv preprint arXiv:2408.02657}, 2024.

\bibitem{han2024infinity}
Jian Han, Jinlai Liu, Yi~Jiang, Bin Yan, Yuqi Zhang, Zehuan Yuan, Bingyue Peng, and Xiaobing Liu.
\newblock Infinity: Scaling bitwise autoregressive modeling for high-resolution image synthesis.
\newblock {\em arXiv preprint arXiv:2412.04431}, 2024.

\bibitem{liu2024visual}
Haotian Liu, Chunyuan Li, Qingyang Wu, and Yong~Jae Lee.
\newblock Visual instruction tuning.
\newblock {\em Advances in neural information processing systems}, 36, 2024.

\bibitem{chen2024sharegpt4v}
Lin Chen, Jinsong Li, Xiaoyi Dong, Pan Zhang, Conghui He, Jiaqi Wang, Feng Zhao, and Dahua Lin.
\newblock Sharegpt4v: Improving large multi-modal models with better captions.
\newblock In {\em European Conference on Computer Vision}, pages 370--387. Springer, 2024.

\bibitem{chen2024internvl}
Zhe Chen, Jiannan Wu, Wenhai Wang, Weijie Su, Guo Chen, Sen Xing, Muyan Zhong, Qinglong Zhang, Xizhou Zhu, Lewei Lu, et~al.
\newblock Internvl: Scaling up vision foundation models and aligning for generic visual-linguistic tasks.
\newblock In {\em Proceedings of the IEEE/CVF Conference on Computer Vision and Pattern Recognition}, pages 24185--24198, 2024.

\bibitem{liu2024sphinx}
Dongyang Liu, Renrui Zhang, Longtian Qiu, Siyuan Huang, Weifeng Lin, Shitian Zhao, Shijie Geng, Ziyi Lin, Peng Jin, Kaipeng Zhang, et~al.
\newblock Sphinx-x: Scaling data and parameters for a family of multi-modal large language models.
\newblock {\em arXiv preprint arXiv:2402.05935}, 2024.

\bibitem{liu2023visual}
Haotian Liu, Chunyuan Li, Qingyang Wu, and Yong~Jae Lee.
\newblock Visual instruction tuning.
\newblock {\em Advances in neural information processing systems}, 36:34892--34916, 2023.

\bibitem{wang2023cogvlm}
Weihan Wang, Qingsong Lv, Wenmeng Yu, Wenyi Hong, Ji~Qi, Yan Wang, Junhui Ji, Zhuoyi Yang, Lei Zhao, Xixuan Song, et~al.
\newblock Cogvlm: Visual expert for pretrained language models.
\newblock {\em arXiv preprint arXiv:2311.03079}, 2023.

\bibitem{bai2023qwen}
Jinze Bai, Shuai Bai, Shusheng Yang, Shijie Wang, Sinan Tan, Peng Wang, Junyang Lin, Chang Zhou, and Jingren Zhou.
\newblock Qwen-vl: A versatile vision-language model for understanding, localization.
\newblock {\em Text Reading, and Beyond}, 2, 2023.

\bibitem{Qwen2.5-VL}
Shuai Bai, Keqin Chen, Xuejing Liu, Jialin Wang, Wenbin Ge, Sibo Song, Kai Dang, Peng Wang, Shijie Wang, Jun Tang, Humen Zhong, Yuanzhi Zhu, Mingkun Yang, Zhaohai Li, Jianqiang Wan, Pengfei Wang, Wei Ding, Zheren Fu, Yiheng Xu, Jiabo Ye, Xi~Zhang, Tianbao Xie, Zesen Cheng, Hang Zhang, Zhibo Yang, Haiyang Xu, and Junyang Lin.
\newblock Qwen2.5-vl technical report.
\newblock {\em arXiv preprint arXiv:2502.13923}, 2025.

\bibitem{Qwen2-VL}
Peng Wang, Shuai Bai, Sinan Tan, Shijie Wang, Zhihao Fan, Jinze Bai, Keqin Chen, Xuejing Liu, Jialin Wang, Wenbin Ge, Yang Fan, Kai Dang, Mengfei Du, Xuancheng Ren, Rui Men, Dayiheng Liu, Chang Zhou, Jingren Zhou, and Junyang Lin.
\newblock Qwen2-vl: Enhancing vision-language model's perception of the world at any resolution.
\newblock {\em arXiv preprint arXiv:2409.12191}, 2024.

\bibitem{dehghani2023scaling}
Mostafa Dehghani, Josip Djolonga, Basil Mustafa, Piotr Padlewski, Jonathan Heek, Justin Gilmer, Andreas~Peter Steiner, Mathilde Caron, Robert Geirhos, Ibrahim Alabdulmohsin, et~al.
\newblock Scaling vision transformers to 22 billion parameters.
\newblock In {\em International Conference on Machine Learning}, pages 7480--7512. PMLR, 2023.

\bibitem{su2021roformer}
Jianlin Su, Yu~Lu, Shengfeng Pan, Ahmed Murtadha, Bo~Wen, and Yunfeng Liu.
\newblock Roformer: enhanced transformer with rotary position embedding. arxiv.
\newblock {\em arXiv preprint arXiv:2104.09864}, 2021.

\bibitem{achiam2023gpt}
Josh Achiam, Steven Adler, Sandhini Agarwal, Lama Ahmad, Ilge Akkaya, Florencia~Leoni Aleman, Diogo Almeida, Janko Altenschmidt, Sam Altman, Shyamal Anadkat, et~al.
\newblock Gpt-4 technical report.
\newblock {\em arXiv preprint arXiv:2303.08774}, 2023.

\bibitem{team2023gemini}
Gemini Team, Rohan Anil, Sebastian Borgeaud, Jean-Baptiste Alayrac, Jiahui Yu, Radu Soricut, Johan Schalkwyk, Andrew~M Dai, Anja Hauth, Katie Millican, et~al.
\newblock Gemini: a family of highly capable multimodal models.
\newblock {\em arXiv preprint arXiv:2312.11805}, 2023.

\bibitem{kong2024hunyuanvideo}
Weijie Kong, Qi~Tian, Zijian Zhang, Rox Min, Zuozhuo Dai, Jin Zhou, Jiangfeng Xiong, Xin Li, Bo~Wu, Jianwei Zhang, et~al.
\newblock Hunyuanvideo: A systematic framework for large video generative models.
\newblock {\em arXiv preprint arXiv:2412.03603}, 2024.

\bibitem{qwen2vl}
Peng Wang, Shuai Bai, Sinan Tan, Shijie Wang, Zhihao Fan, Jinze Bai, Keqin Chen, Xuejing Liu, Jialin Wang, Wenbin Ge, et~al.
\newblock Qwen2-vl: Enhancing vision-language model's perception of the world at any resolution.
\newblock {\em arXiv preprint arXiv:2409.12191}, 2024.

\bibitem{zhang2023internlm}
Pan Zhang, Xiaoyi Dong, Bin Wang, Yuhang Cao, Chao Xu, Linke Ouyang, Zhiyuan Zhao, Haodong Duan, Songyang Zhang, Shuangrui Ding, et~al.
\newblock Internlm-xcomposer: A vision-language large model for advanced text-image comprehension and composition.
\newblock {\em arXiv preprint arXiv:2309.15112}, 2023.

\bibitem{zhang2024towards}
Kaiyan Zhang, Biqing Qi, and Bowen Zhou.
\newblock Towards building specialized generalist ai with system 1 and system 2 fusion.
\newblock {\em arXiv preprint arXiv:2407.08642}, 2024.

\bibitem{xiao2024florence}
Bin Xiao, Haiping Wu, Weijian Xu, Xiyang Dai, Houdong Hu, Yumao Lu, Michael Zeng, Ce~Liu, and Lu~Yuan.
\newblock Florence-2: Advancing a unified representation for a variety of vision tasks.
\newblock In {\em Proceedings of the IEEE/CVF Conference on Computer Vision and Pattern Recognition}, pages 4818--4829, 2024.

\bibitem{geva2020transformer}
Mor Geva, Roei Schuster, Jonathan Berant, and Omer Levy.
\newblock Transformer feed-forward layers are key-value memories.
\newblock {\em arXiv preprint arXiv:2012.14913}, 2020.

\bibitem{zhang2021tip}
Renrui Zhang, Rongyao Fang, Wei Zhang, Peng Gao, Kunchang Li, Jifeng Dai, Yu~Qiao, and Hongsheng Li.
\newblock Tip-adapter: Training-free clip-adapter for better vision-language modeling.
\newblock {\em arXiv preprint arXiv:2111.03930}, 2021.

\bibitem{sukhbaatar2019augmenting}
Sainbayar Sukhbaatar, Edouard Grave, Guillaume Lample, Herve Jegou, and Armand Joulin.
\newblock Augmenting self-attention with persistent memory.
\newblock {\em arXiv preprint arXiv:1907.01470}, 2019.

\bibitem{ma2025inference}
Nanye Ma, Shangyuan Tong, Haolin Jia, Hexiang Hu, Yu-Chuan Su, Mingda Zhang, Xuan Yang, Yandong Li, Tommi Jaakkola, Xuhui Jia, et~al.
\newblock Inference-time scaling for diffusion models beyond scaling denoising steps.
\newblock {\em arXiv preprint arXiv:2501.09732}, 2025.

\bibitem{xie2025sana}
Enze Xie, Junsong Chen, Yuyang Zhao, Jincheng Yu, Ligeng Zhu, Yujun Lin, Zhekai Zhang, Muyang Li, Junyu Chen, Han Cai, et~al.
\newblock Sana 1.5: Efficient scaling of training-time and inference-time compute in linear diffusion transformer.
\newblock {\em arXiv preprint arXiv:2501.18427}, 2025.

\bibitem{snell2024scaling}
Charlie Snell, Jaehoon Lee, Kelvin Xu, and Aviral Kumar.
\newblock Scaling llm test-time compute optimally can be more effective than scaling model parameters.
\newblock {\em arXiv preprint arXiv:2408.03314}, 2024.

\bibitem{ho2022classifier}
Jonathan Ho and Tim Salimans.
\newblock Classifier-free diffusion guidance.
\newblock {\em Advances in Neural Information Processing Systems Workshops (NeurIPS Workshops)}, 2021.

\bibitem{lu2022dpm}
Cheng Lu, Yuhao Zhou, Fan Bao, Jianfei Chen, Chongxuan Li, and Jun Zhu.
\newblock Dpm-solver++: Fast solver for guided sampling of diffusion probabilistic models.
\newblock {\em arXiv preprint arXiv:2211.01095}, 2022.

\bibitem{kirstain2023pick}
Yuval Kirstain, Adam Polyak, Uriel Singer, Shahbuland Matiana, Joe Penna, and Omer Levy.
\newblock Pick-a-pic: An open dataset of user preferences for text-to-image generation.
\newblock {\em Advances in Neural Information Processing Systems}, 36:36652--36663, 2023.

\bibitem{zhang2019root}
Biao Zhang and Rico Sennrich.
\newblock Root mean square layer normalization.
\newblock {\em Advances in Neural Information Processing Systems}, 32, 2019.

\bibitem{loshchilov2017decoupled}
I~Loshchilov.
\newblock Decoupled weight decay regularization.
\newblock {\em arXiv preprint arXiv:1711.05101}, 2017.

\bibitem{ma2024janusflow}
Yiyang Ma, Xingchao Liu, Xiaokang Chen, Wen Liu, Chengyue Wu, Zhiyu Wu, Zizheng Pan, Zhenda Xie, Haowei Zhang, Liang Zhao, et~al.
\newblock Janusflow: Harmonizing autoregression and rectified flow for unified multimodal understanding and generation.
\newblock {\em arXiv preprint arXiv:2411.07975}, 2024.

\bibitem{baldridge2024imagen}
Jason Baldridge, Jakob Bauer, Mukul Bhutani, Nicole Brichtova, Andrew Bunner, Kelvin Chan, Yichang Chen, Sander Dieleman, Yuqing Du, Zach Eaton-Rosen, et~al.
\newblock Imagen 3.
\newblock {\em arXiv preprint arXiv:2408.07009}, 2024.

\bibitem{elo1978rating}
Arpad~E Elo and Sam Sloan.
\newblock The rating of chessplayers: Past and present.
\newblock {\em Arco Pub}, 1978.

\bibitem{kolors2024}
Kuaishou Technology.
\newblock Kolors.
\newblock \url{https://github.com/Kwai-Kolors/Kolors}, 2024.

\bibitem{saharia2022photorealistic}
Chitwan Saharia, William Chan, Saurabh Saxena, Lala Li, Jay Whang, Emily~L Denton, Kamyar Ghasemipour, Raphael Gontijo~Lopes, Burcu Karagol~Ayan, Tim Salimans, et~al.
\newblock Photorealistic text-to-image diffusion models with deep language understanding.
\newblock {\em Advances in neural information processing systems}, 35:36479--36494, 2022.

\bibitem{team2024gemma}
Gemma Team, Morgane Riviere, Shreya Pathak, Pier~Giuseppe Sessa, Cassidy Hardin, Surya Bhupatiraju, L{\'e}onard Hussenot, Thomas Mesnard, Bobak Shahriari, Alexandre Ram{\'e}, et~al.
\newblock Gemma 2: Improving open language models at a practical size.
\newblock {\em arXiv preprint arXiv:2408.00118}, 2024.

\end{thebibliography}

\end{document}